\def\AREALSNAMED{1}
\documentclass{article}

\usepackage{microtype}
\usepackage{graphicx}
\usepackage{booktabs}
\usepackage{array}
\usepackage{float}
\usepackage{hyperref}
\newif\ifblind
\ifdefined\AREALSNAMED\blindfalse\else\blindtrue\fi
\ifblind
  \IfFileExists{icml2026.sty}{\usepackage{icml2026}}{\usepackage{icml2026_stub}}
\else
  \usepackage[preprint]{icml2026_stub}
\fi

\usepackage{amsmath}
\usepackage{amssymb}
\usepackage{mathtools}
\usepackage{amsthm}
\theoremstyle{plain}
\newtheorem{proposition}{Proposition}
\newtheorem{lemma}{Lemma}
\usepackage{xcolor}
\usepackage{tikz}
\usetikzlibrary{arrows.meta,calc}
\usepackage{pgfplots}
\pgfplotsset{compat=1.18}
\usepgfplotslibrary{groupplots,fillbetween}
\usepackage[capitalize,noabbrev]{cleveref}
\makeatletter\providecommand\captionof[1]{\def\@captype{#1}\caption}\makeatother

\definecolor{arealteal}{HTML}{0A7A72}
\definecolor{arealwarn}{HTML}{B4541A}
\definecolor{arealmuted}{HTML}{5B676C}
\definecolor{arealfg}{HTML}{1B2326}
\definecolor{arealgrid}{HTML}{E3E7E8}
\definecolor{arealrec}{HTML}{8AC4BB}
\hypersetup{colorlinks=true,linkcolor=arealteal,citecolor=arealteal,urlcolor=arealteal}
\hypersetup{pdftitle={Large Geometry Models Are Priors for Fast Physics Pretraining}}
\ifdefined\AREALSNAMED\hypersetup{pdfauthor={Maggie Chlon and Leon Chlon}}\fi

\newcommand{\nMeshes}{61{,}822}
\newcommand{\nParams}{3.2M}
\newcommand{\gapClosed}{90.7\%}
\newcommand{\gapCI}{84.4--96.9\%}
\newcommand{\heldGap}{91.7\%}

\newcommand{\decoderGain}{11.4\%}
\newcommand{\pointMaeGain}{6.1\%}
\newcommand{\densityReduction}{81.5\%}

\newcommand{\areals}{AREALS}
\newcommand{\E}{\mathbb{E}}
\newcommand{\nnT}{nn^{\top}}

\icmltitlerunning{Large Geometry Models Are Priors for Fast Physics Pretraining}

\def\authoraffiliation{Hassana Labs}
\def\authoremail{lc574@cantab.ac.uk}
\ifblind\else\ifx\authoraffiliation\empty\renewcommand{\icmlauthor}[2]{{\bfseries #1}}\fi\fi

\begin{document}

\twocolumn[
\icmltitle{Large Geometry Models Are Priors for Fast Physics Pretraining}

\icmlsetsymbol{equal}{*}
\begin{icmlauthorlist}
\icmlauthor{Maggie Chlon}{aff}
\icmlauthor{Leon Chlon}{aff}
\end{icmlauthorlist}
\icmlaffiliation{aff}{\authoraffiliation}
\icmlcorrespondingauthor{Leon Chlon}{\authoremail}
\icmlkeywords{masked autoencoders, inverse problems, geometric deep learning, neural operators, CFD}

\vskip 0.10in
\begin{center}
\pgfplotstableread[row sep=\\]{%
  budget scratch scratchlo scratchhi areals arealslo arealshi \\
  18 0.44783 0.41559 0.48301 0.27479 0.24758 0.30220 \\
  35 0.34900 0.31890 0.37842 0.22266 0.20029 0.24510 \\
  69 0.26989 0.24185 0.29744 0.19246 0.16986 0.21774 \\
  137 0.21173 0.18988 0.23635 0.16570 0.14315 0.19238 \\
  274 0.18732 0.16331 0.21402 0.14605 0.12587 0.16905 \\
}\arealsEffWindsor
\pgfplotstableread[row sep=\\]{%
  budget scratch scratchlo scratchhi areals arealslo arealshi \\
  25 0.48944 0.47332 0.50561 0.36870 0.36031 0.37704 \\
  50 0.41597 0.40183 0.43079 0.30812 0.30015 0.31558 \\
  100 0.34679 0.33244 0.36203 0.25339 0.24663 0.26034 \\
  200 0.26909 0.25913 0.27878 0.20937 0.20468 0.21430 \\
  400 0.21935 0.21323 0.22574 0.18089 0.17685 0.18511 \\
}\arealsEffDrivAerML
\pgfplotstableread[row sep=\\]{%
  budget scratch scratchlo scratchhi areals arealslo arealshi \\
  406 0.30346 0.29783 0.30939 0.27712 0.27291 0.28139 \\
  812 0.27650 0.27048 0.28273 0.25493 0.25168 0.25843 \\
  1623 0.25637 0.25215 0.26061 0.23860 0.23568 0.24184 \\
  3246 0.24746 0.24421 0.25091 0.23257 0.22965 0.23581 \\
  6492 0.24603 0.24193 0.25052 0.23026 0.22742 0.23341 \\
}\arealsEffDrivAerNetPP
\definecolor{arealscratch}{HTML}{7A5195}%
\def\arealsEffPanel#1#2#3#4{
  \addplot[name path=sl, draw=none, forget plot] table[x=budget, y=scratchlo] {#1};
  \addplot[name path=sh, draw=none, forget plot] table[x=budget, y=scratchhi] {#1};
  \addplot[arealscratch, fill opacity=0.15, draw=none, forget plot] fill between[of=sl and sh];
  \addplot[name path=al, draw=none, forget plot] table[x=budget, y=arealslo] {#1};
  \addplot[name path=ah, draw=none, forget plot] table[x=budget, y=arealshi] {#1};
  \addplot[arealteal, fill opacity=0.15, draw=none, forget plot] fill between[of=al and ah];
  \addplot[arealscratch, line width=1.2pt, mark=*, mark size=1.9pt, mark options={fill=white, line width=0.9pt}]
    table[x=budget, y=scratch] {#1};
  \addplot[arealteal, line width=1.4pt, mark=*, mark size=2.1pt, mark options={draw=white, line width=0.4pt}]
    table[x=budget, y=areals] {#1};
  \draw[-{Stealth[length=4.5pt,width=4pt]}, arealfg, line width=0.8pt] (axis cs:#3,0.25) -- (axis cs:#2,0.25);
  \node[anchor=north east, font=\scriptsize, text=arealfg, align=right] at (rel axis cs:0.97,0.97)
    {#4\% fewer\\simulations};
}%
\begin{tikzpicture}
\begin{groupplot}[
    group style={group size=3 by 1, horizontal sep=1.05cm},
    width=0.355\textwidth, height=4.4cm,
    xmode=log, log basis x=2, xminorticks=false,
    enlarge x limits=0.07,
    xlabel={Training simulations},
    tick label style={font=\scriptsize},
    label style={font=\small},
    title style={font=\small, yshift=-3pt},
    axis line style={arealmuted}, tick style={arealmuted},
    ymajorgrids, grid style={arealgrid},
    extra y ticks={0.25}, extra y tick labels={},
    extra y tick style={grid=major, major grid style={arealmuted, dashed, line width=0.7pt}},
    yticklabel style={/pgf/number format/fixed, /pgf/number format/fixed zerofill, /pgf/number format/precision=2},
  ]
  \nextgroupplot[title={WindsorML}, ylabel={Test nRMSE, no sensors},
    ymin=0.11, ymax=0.50, ytick={0.2,0.3,0.4},
    xtick={18,35,69,137,274}, xticklabels={18,35,69,137,274},
    legend to name=arealsefflegend, legend columns=3,
    legend style={font=\scriptsize, draw=none, column sep=10pt}]
  \arealsEffPanel{\arealsEffWindsor}{35}{137}{74.5}
  \addlegendentry{from scratch}
  \addlegendentry{\areals{}}
  \addlegendimage{arealmuted, dashed, line width=0.7pt}
  \addlegendentry{target, nRMSE 0.25}
  \nextgroupplot[title={DrivAerML}, ymin=0.16, ymax=0.52, ytick={0.2,0.3,0.4,0.5},
    xtick={25,50,100,200,400}, xticklabels={25,50,100,200,400}]
  \arealsEffPanel{\arealsEffDrivAerML}{200}{400}{50}
  \nextgroupplot[title={DrivAerNet++}, ymin=0.22, ymax=0.315, ytick={0.22,0.24,0.26,0.28,0.30},
    xtick={406,812,1623,3246,6492}, xticklabels={406,812,{1,623},{3,246},{6,492}}]
  \arealsEffPanel{\arealsEffDrivAerNetPP}{1623}{3246}{50}
\end{groupplot}
\node[anchor=south] at ([yshift=13pt]group c2r1.north) {\pgfplotslegendfromname{arealsefflegend}};
\end{tikzpicture}

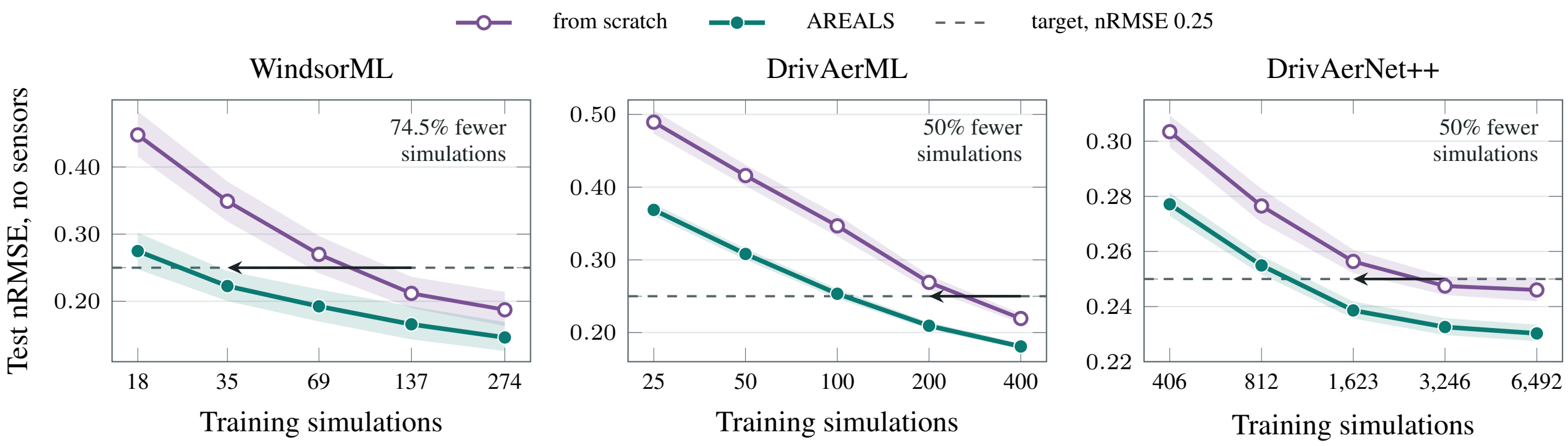
\captionof{figure}{\textbf{Euler--Lagrange geometry pretraining reaches the same accuracy with half the simulations or fewer.} Test nRMSE (normalised RMSE) without sensors against training simulations, at 8,000 updates per fit; lines are means
over three seeds with pointwise 95\% bootstrap bands, and vertical scales differ. Arrows join the smallest budgets at
which each model reaches the 0.25 target. At every budget, the paired 95\% interval of the difference excludes zero.}
\label{fig:efficiency}
\end{center}
\vskip 0.12in
]

\ifblind\printAffiliationsAndNotice{}\else\ifx\authoraffiliation\empty\else\printAffiliationsAndNotice{}\fi\fi


\begin{abstract}
Language and vision models became data-efficient by pretraining on plentiful unlabelled data; physical field models
have such data, over ten million 3D shapes, but geometry pretraining designs its targets rather than deriving them from
the physics. Euler--Lagrange geometry pretraining derives them: for attached flow, a shape enters the flow's
Euler--Lagrange equation only through coefficients that distance and normal line (the normal up to sign) fix to first order. Their reconstruction error therefore bounds what a representation loses about any field, and
fine-tuning on costly simulations need only assemble the global field. We realise it in \areals{}, a masked surface autoencoder whose
\emph{areal tokens} carry position, normal and area, pretrained self-supervised and sign-free on \nMeshes{} raw
meshes. Because the operating condition enters only through the far field, sensors can supply
it: \areals{} completes a hidden field in one forward pass without condition labels. It matches the accuracy of the same
architecture trained from scratch with half the simulations on DrivAerML and DrivAerNet++ and about a quarter on
WindsorML. On DoubleDelta wings, 20 sensors give 3.9 and 4.0 times lower field and force error than GABI, a Bayesian
inversion method with a learned prior, and beyond the training angles pretraining halves the error.
\end{abstract}

\section{Introduction}
\label{sec:intro}

Pretraining on plentiful unlabelled data is what made language and vision models data-efficient
\citep{devlin2019bert,he2022mae}. Physical field models have the plentiful data, with over ten million 3D shapes in
Objaverse-XL alone \citep{deitke2023objaversexl}, but not the pretraining: what to learn from geometry has been designed
rather than derived from the physics \citep{geopt2026,zhang2026frozen}. The closest attempt, GeoPT, abandons geometry
alone: in its own tests, pretraining to predict vector distance degraded accuracy \citep{geopt2026}. Its network
instead takes synthetic velocity fields as input in pretraining and the freestream velocity in fine-tuning, so it must
be told the operating condition. Designed targets carry no guarantee, because
a shape can be reproduced closely while its field is not: a wall can converge to a flat one in position while its
pressure does not (\cref{sec:bounds}). Field models therefore still learn geometry from simulations that cost hours
each \citep{li2023gino,wu2024transolver,alkin2025abupt}.

Euler--Lagrange geometry pretraining is, to our knowledge, the first to take its targets from the governing equation,
learning from meshes alone the geometry through which that equation acts. Geometry thereby becomes a prior for
field models: it supplies the field's local information, and fine-tuning on simulations need only assemble the global
field at any query point, cutting the simulations required to learn the same physics. Realised in \areals{}, a masked
surface autoencoder, it needs half the simulations of the same architecture trained from scratch to reach the same
accuracy on DrivAerML \citep{ashton2024drivaerml} and DrivAerNet++ \citep{drivaernetpp} and about a quarter on WindsorML
\citep{ashton2024windsorml} (\cref{fig:efficiency}).

\begin{figure*}[t]
\centering
\includegraphics[width=\textwidth]{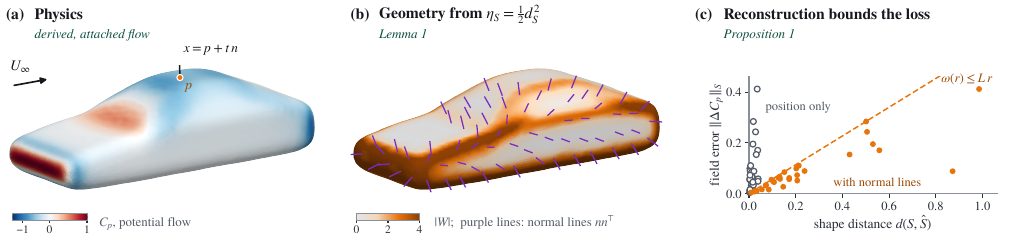}
\caption{\textbf{What the flow equation needs from geometry, computed on one car mesh in potential flow}
(\cref{app:formal}). Tags say how each panel is established: derived for attached flow, or proved. (a)~Surface
pressure; a point $x$ near the wall is its foot $p$ plus a distance $t$ along the normal. (b)~The Hessian of the half
squared distance $\eta_S$ gives the normal lines (purple), and its third derivative the shape operator $W$, whose norm
sets the colour. (c)~Field error against shape distance for 30 perturbed shapes, with the flow re-solved on each: per
unit distance it grows from 1.8 to 24 as the perturbing waves shorten when the distance counts position alone (open),
but stays at 0.45--0.57 when it also counts normal lines (filled), so $\omega(r)\le0.57\,r$. \Cref{fig:wallcoords}
draws the wall coordinates and $\eta_S$ on the centre plane.}
\label{fig:hero}
\end{figure*}

The governing equation takes only local information from geometry. Wall coordinates locate a point by its nearest point on the surface, its foot, and its distance from it. Written in
them, the equation near the surface depends on the body
only through the surface metric and shape operator at the foot, which measure lengths and curvature on the
surface, and their derivatives. All of these are read from one function, the half squared distance to the surface: its value gives the surface, its
second derivative the normal line (the surface normal up to sign), and its third derivative the shape operator
(\cref{lem:primitive,fig:hero}). For attached, irrotational flow this is exact: the governing equation is the
Euler--Lagrange equation of an action whose density is the pressure \citep{bateman1929variational}, and a nearby shape
changes only the coefficients of one fixed equation on one fixed domain (\cref{sec:el}). Distance and normal line fix
those coefficients to first order and curvature enters only through their derivatives, so distance and normal line are
the pretraining targets: reconstruction sets up the flow equation, and simulations solve it.

For the separated, turbulent flows we test, the argument motivates the targets rather than guaranteeing them. Its
predictions still hold in a matched comparison (\cref{sec:exp-targets}), and turbulence closures such as Spalart--Allmaras and $k$--$\omega$ shear-stress transport take the wall distance itself as an input
\citep{spalart1994oneequation,menter1994twoequation}.

What remains is the operating condition, which on a physical prototype often goes unrecorded. It enters the flow only
through the far field, so a handful of sensors can supply it. \areals{} reads a hidden field from sparse sensors in one
forward pass without ever observing the condition, the first sparse-sensor field model pretrained on raw, unlabelled
meshes. Our contributions are:
\begin{itemize}
\setlength{\itemsep}{1pt}\setlength{\topsep}{2pt}\setlength{\parskip}{0pt}
\item \textbf{\areals{}}, which realises the method as a masked surface autoencoder: areal tokens carry position,
normal and area and weight attention by area, and sign-free targets admit \nMeshes{} raw meshes without repair
(\cref{sec:method}).
\item \textbf{Half the simulations to learn the same field.} Pretraining reaches a target accuracy with 50--74.5\% fewer
simulations than the same architecture from scratch, and at matched data the derived targets cut error by 22--40\%,
while curvature adds nothing and local coefficients alone transfer worse than scratch
(\cref{sec:exp-pretrain,sec:exp-targets}).
\item \textbf{Sensors in place of the operating condition.} 20 sensors close 84\% of the gap to a model told the true
angle, with 3.9 times lower error than GABI \citep{vadeboncoeur2026gabi}, and outside the training angles, where the
model from scratch stops reading its sensors, pretraining halves its error (\cref{sec:exp-sensors}).
\item \textbf{Guarantees for the derived targets.} Reconstructing distance and normal line to Hausdorff error $d$ bounds the best field predictor's mean squared error
by $\E\,\omega(2d)^2$, where $\omega$, the field's modulus of continuity in shape, is $O(r)$ in linearised potential
flow; both targets are necessary (\cref{prop:sufficiency,prop:outer}).
\item \textbf{Mechanistic evidence.} Pretrained attention concentrates within a quarter of the surface radius, and
fine-tuning extends it on every test wing (\cref{sec:exp-mechanism}); transfer runs through the query decoder that masked pretraining usually discards, with
which the \areals{} encoder beats a Point-MAE encoder \citep{pang2022pointmae} nearly ten times its size.
\end{itemize}
\section{Euler--Lagrange Geometry Pretraining}
\label{sec:inversion}

Let $S$ be a surface with area element $dA$ and $f$ a field on it, such as pressure. The field is set by the shape and
an operating condition $c$, such as a flow angle: $f=F(S,c)$, where $F$ is the solution operator of the governing
equations, and $c$ may go unreported. Euler--Lagrange geometry pretraining learns from $S$ alone what $F$ needs from it
(\cref{sec:el,sec:bounds}); simulations and sensors supply the rest (\cref{sec:levels}).

\subsection{The flow equation sees geometry locally}
\label{sec:el}

For attached, irrotational flow, the governing equation is an Euler--Lagrange equation, and it is local. In a
barotropic fluid with velocity $\nabla\varphi$, locality, the shift symmetry $\varphi\mapsto\varphi+\mathrm{const}$,
rotations and Galilean boosts fix the Lagrangian density as the pressure $p$, up to the equation of state
\citep{bateman1929variational,seliger1968variational,son2006general}. Stationarity of the action
$\int\!\!\int p\,d^3x\,d\tau$ over the fluid and time $\tau$ gives conservation of mass, with no flow through the wall,
$\partial_n\varphi=0$, as a natural boundary condition (\cref{app:formal,fig:guarantees}).

Near the wall, write a point as $x=\pi_S(x)+t\,n$, with $\pi_S(x)$ its foot on $S$, $n$ the normal there and $t$ the
signed distance (\cref{fig:hero}a). The Euclidean metric becomes $g=g_S\big((I+tW)\,\cdot,(I+tW)\,\cdot\big)+dt^2$,
with $g_S$ the surface metric and $W$ the shape operator at the foot. Within the reach $\ell$ of $S$, the distance up to
which every point has a unique foot, the action therefore sees the shape only through $g_S$ and $W$. A nearby shape
$S'$ alters only coefficients, not the equation or its domain: carried back by a map $T$ that moves $S$ along its
normals onto $S'$, the potential $\tilde\varphi$ of incompressible flow around $S'$ solves
\begin{equation}
\nabla\cdot\big(A\nabla\tilde\varphi\big)=0,\qquad A=\sqrt{\det g_T}\;g_T^{-1},
\label{eq:coefficients}
\end{equation}
with $A\nabla\tilde\varphi\cdot n=0$ on $S$ and $g_T$ the metric that $T$ pulls back. Errors $\varepsilon$ in position
and $\delta$ in normal line move $A$ by at most a constant times $\varepsilon/\ell+\delta$, and an error in curvature
enters only through the derivatives of $A$ (\cref{app:formal}).

Neither $g_S$ nor $W$ depends on how $S$ is sampled, and reversing the orientation, $(n,t,W)\mapsto(-n,-t,-W)$, leaves
$g$ unchanged, so the action is blind to sampling and orientation. So is the half squared distance
$\eta_S=\tfrac12d_S^2$ to $S$, which exists for every mesh and from which all of this geometry is read
(\cref{fig:hero}b):

\begin{lemma}[Distance determines geometry]
\label{lem:primitive}
For every closed set $S\subset\mathbb{R}^3$, $\tfrac12|x|^2-\eta_S$ is convex and $S=\eta_S^{-1}(0)$. Where $S$ is $C^3$,
$\nabla\eta_S(x)=x-\pi_S(x)$ for the closest-point map $\pi_S$, $\nabla^2\eta_S=\nnT$ on $S$, and
$(\partial_X\nabla^2\eta_S)\,n=\partial_Xn$ for tangent $X$: value, Hessian and third derivative give the surface, its
normal line and its shape operator, which set the metric of the wall coordinates $(\pi_S,d_S)$. If
$|\hat d-d_S|\le\varepsilon$, then $\{d_S\le r-\varepsilon\}\subseteq\{\hat d\le r\}\subseteq\{d_S\le r+\varepsilon\}$
for all $r$, and the $k$-th derivative of $\hat d-d_S$ smoothed at width $r$ is at most $C_k\varepsilon/r^k$.
\end{lemma}

\subsection{Reconstruction bounds what the representation loses}
\label{sec:bounds}

By \cref{lem:primitive}, a learned distance within $\varepsilon$ of $d_S$ fixes position and topology at resolution
$\varepsilon$ \citep{cohensteiner2007stability}, but gives normals only to $\varepsilon/r$ at width $r$ and curvature to
$\varepsilon/r^2$, and position alone does not fix normals \citep{hildebrandt2006convergence}. \areals{} therefore
reconstructs the value and the Hessian of $\eta_S$ directly: the distance and the projector $\nnT$ onto the normal line,
which stays continuous where $\nabla d_S$ flips sign. Curvature, one derivative higher, is not supervised.

A representation that pins down the surface and its normal lines pins down every field, up to how strongly the field
responds to a change of shape. Let $\hat S=\mathrm{dec}(\mathrm{enc}(S))$ be the surface, with its normal lines,
rebuilt from the representation of a fully visible $S$, and $d$ the Hausdorff distance between the sets
$\{(x,\nnT(x))\}$ of two surfaces.

\begin{proposition}[Reconstruction bounds what the representation loses]
\label{prop:sufficiency}
Let $\omega(r)$ be the supremum of $\lVert F(S,c)-F(S',c)\circ\pi_{S'}\rVert_S$ over conditions $c$ and admissible
shapes with $d(S,S')\le r$, where $\pi_{S'}$ is a measurable closest-point map from $S$ to $S'$, $F(S',c)$ is continuous
on $S'$ and $\lVert\cdot\rVert_S$ is the area-weighted root-mean-square norm on $S$. Then the best predictor of $f(x)$
from the representation $\mathrm{enc}(S)$, the condition and the query point $x$, drawn from the area measure of $S$,
has mean squared error at most $\E\,\omega\big(2d(S,\hat S)\big)^2$.
\end{proposition}

No stability of $F$ is assumed (\cref{app:formal}): exact reconstruction loses nothing, and $\omega(r)\le Lr$ gives
$4L^2\,\E\,d(S,\hat S)^2$. Both targets are also necessary: the distance to $S$ from $x+v$ is itself a field, equal to
$|v\cdot n(x)|$ to first order in $v$, so predicting every such field requires both. The bound concerns the best
predictor, not whether a finite network learns the long-range operator from scarce simulations, which
\cref{sec:experiments} measures.

Position alone gives no such bound. In linearised subsonic flow a wall $y=h\cos kx$ converges to a flat one in
Hausdorff distance as $h\to0$ with $hk$ fixed, yet its pressure coefficient $-(2hk/\beta)\cos kx$, with
$\beta=\sqrt{1-M^2}$, does not; with normal lines in the metric, $\omega(r)$ is of order $r/\beta$. More generally,
around a smooth body in potential flow the wall pressure changes at most linearly with errors in position, normal line
and curvature (\cref{prop:outer} in \cref{app:formal}). In linearised flow the curvature term drops out, so there
$\omega(r)=O(r)$, as computed on a car mesh in \cref{fig:hero}c. For separated flow and the Reynolds-averaged
Navier--Stokes (RANS) fields of \cref{sec:experiments}, $\omega$ is not known; there, the matched comparison of
\cref{sec:exp-targets} tests the targets directly.

\subsection{What simulations and sensors add}
\label{sec:levels}

\paragraph{Long-range dependence.} Local geometry does not fix the whole field: in linearised subsonic flow, for
instance, the whole wall contributes to the pressure gradient at each point (\cref{app:formal}). In potential flow the
long-range part is an integral over the surface against the Green's function $\Gamma(x-y)$ and its normal derivative
\citep{katz2001lowspeed}, kernels that depend on relative position, normal and area, all of which an areal token carries. The area-weighted attention of \cref{eq:attn} has the same form, up to softmax
normalisation. As the targets
are nearest-surface quantities, this suggests a division of work: geometry pretraining shapes short-range kernels,
simulations add the long-range ones, and sensors fix the condition, which enters only through the far field.
\Cref{sec:exp-mechanism} tests this division in the attention of trained models.

\paragraph{Level 1: geometry is the prior over field models.} Inference has two levels: learning field models from
scarce simulations, and reading a new field from sparse sensors. The first interprets fine-tuning: geometry
pretraining fixes the centre $\theta_{\mathrm{geo}}$ of a Gaussian prior over the weights of a field model before any
simulation is seen. Fine-tuning with a pull towards $\theta_{\mathrm{geo}}$ targets the posterior
mode given the simulations \citep{li2018l2sp,shwartzziv2022pretrain}, and starting from $\theta_{\mathrm{geo}}$ without
it, as in our runs, does so approximately \citep{ali2019continuous}. The prior is learned from other meshes without any
field, so it is chosen independently of the simulations, as PAC-Bayes bounds require (\cref{app:formal}).

\paragraph{Level 2: sensors are evidence about the condition.} On a new shape, a mask splits $S$ into a visible part
$V$ and a hidden part $H=S\setminus V$: hidden patches, or everything except $K$ sensor points. We observe $y=f|_V$ and
seek $f_H=f|_H$. The shape fixes everything except the condition, so the hidden values depend on the sensors only
through it:
\begin{equation}
\begin{aligned}
p(c\mid y,S)&\propto p(c)\,p(y\mid S,c),\\
p(f_H\mid y,S)&=\int p(f_H\mid S,c)\,p(c\mid y,S)\,dc .
\end{aligned}
\label{eq:posterior}
\end{equation}
Under area-weighted squared error, the Bayes estimator of the hidden field is the posterior mean
$\hat f(x)=\E[f(x)\mid y,S]$, $x\in H$ \citep{stuart2010inverse}. \areals{} amortises it: its sensor model is trained with this loss on masked
simulations, so its target is the posterior mean under the simulations' distribution of shapes, conditions and masks
\citep{adler2018deep}, and inference is one forward pass. The sensor model never receives the condition, so it can
train on simulations whose conditions are unrecorded or too complex to parametrise, and needs no search over $c$;
evaluating \cref{eq:posterior} explicitly could at best recover the true condition. Beyond the simulations'
distribution, as at the extrapolated angles of \cref{sec:exp-beyond}, the Bayesian reading lapses and accuracy reflects
the network's inductive bias.

\paragraph{What measurements are worth.} Three identically trained networks bracket the value of measurements: the
shape-only model $\E[f\mid S]$, the sensor model $\E[f\mid S,y]$ and the true-condition model $\E[f\mid S,c]$, which is
told the condition instead. From their test errors $\mathcal{E}$, in mean normalised MSE or nRMSE as each result
states, the gap closure $G=\big(\mathcal{E}(S)-\mathcal{E}(S,y)\big)/\big(\mathcal{E}(S)-\mathcal{E}(S,c)\big)$ is the
fraction of the true condition's advantage that the sensors deliver. For exact estimators $0\le G\le1$, since
conditioning never increases the Bayes risk and shape and condition fix the field; measured $G$ also reflects each
network's approximation error.

\section{\areals{}}
\label{sec:method}

\areals{} realises Euler--Lagrange geometry pretraining with one encoder and decoder trained twice, first to complete
hidden geometry and then to complete a hidden field on a known surface (\cref{fig:pretrain}); only the input columns
and the output head change. Each design choice builds in a property of \cref{sec:inversion}. The targets are distance and normal line. The normal
target is sign-free and attention is weighted by area because the action is blind to orientation and to sampling, and
each token carries position, normal and area,
the arguments of the surface-integral kernels of \cref{sec:levels}.

\subsection{Areal tokens}
\label{sec:tokens}

Every attention layer integrates over the surface rather than over the samples. Given surface samples $x_i$ with
normals $n_i$ and area weights $w_i$ that sum to the surface area, it adds the log area weight of each key to its
logit, as proportional attention does for merged tokens \citep{bolya2023tome},
\begin{equation}
\mathrm{Attn}(q)=\frac{\sum_j w_j\,e^{q\cdot k_j}\,v_j}{\sum_j w_j\,e^{q\cdot k_j}}
\approx\frac{\int_S e^{q\cdot k(x)}\,v(x)\,dA(x)}{\int_S e^{q\cdot k(x)}\,dA(x)},
\label{eq:attn}
\end{equation}
so each key counts in proportion to the area it represents, not to how densely its part of the surface was sampled.
Farthest-point sampling selects 256 anchors over the visible surface, or the whole surface when reading fields, and each
keeps its normal and the area it represents. A cross-attention block reads the visible samples into one token per
anchor, self-attention blocks mix the tokens, and rotary embeddings carry 3D position into every attention layer. The
encoder's output for a shape is its 256 tokens with their anchor positions, normals and area weights. The model has \nParams{} parameters
(\cref{app:impl}).

\subsection{Masked geometry pretraining}
\label{sec:pretraining}

Pretraining is self-supervised: both targets of \cref{sec:bounds} are computed from the mesh itself, and the model
reconstructs them from a partly hidden surface. We hide 60\% of the surface
area as 32 patches, and the decoder sends query points anywhere in space, most of them near the surface, to
cross-attend to the tokens. The first target is the unsigned distance to the full surface. The second is the normal line
as the symmetric matrix $\nnT$, trained with the squared Frobenius distance to the target projector and read as its top
eigenvector, which under this loss is a Bayes estimate of the normal line (\cref{app:impl}). A unit-vector head would
have to turn through $180^\circ$ around every free edge; the projector head never turns. Neither target requires a
watertight or consistently oriented mesh, so we pretrain on \nMeshes{} raw Objaverse-XL meshes \citep{deitke2023objaversexl} for 10,626 updates. With 60\% of each surface hidden, \areals{}
reconstructs validation shapes with a distance error of 0.098 in nRMSE and a normal-line error of $22.0^\circ$
(\cref{app:impl}).

\paragraph{Surfel and symmetry readouts.}\label{sec:priors} Read as disks, the tokens reconstruct visible geometry without training.
Rigid symmetries found by transformation-space voting \citep{mitra2006symmetry} mark a hidden
point as observed when its image is visible, cutting normal error on hidden copies of visible parts from $10.9^\circ$ to
$6.9^\circ$ (\cref{fig:priors}, \cref{app:tubes,app:priors}).

\subsection{Field completion from sparse sensors}
\label{sec:fieldmode}

For fields, \areals{} gains two input columns in its first embedding layer, the standardised field value times a
presence flag and the flag itself, both initialised to zero so that fine-tuning starts exactly from the geometry
encoder. A linear head on the decoder's query features predicts the field at any point. The loss is the area-weighted
squared error over query points, weighted by 1 on hidden and 0.25 on visible ones, and its minimiser is the posterior
mean $\E[f\mid y,S]$ of \cref{sec:levels}. Training mixes examples with nothing visible and examples with observations, so the sensor model
also predicts when nothing is observed; for the gap closure, the shape-only model is a separate network trained without
observations.

\section{Experiments}
\label{sec:experiments}

Each subsection tests one claim of \cref{sec:intro}. Unless stated otherwise, every model is fine-tuned from the one
pretrained on all \nMeshes{} meshes (\cref{sec:pretraining}). The matched comparison of \cref{sec:exp-targets} pretrains its own priors, and the decoder and design tests of \cref{sec:exp-decoder,sec:exp-design} use priors
pretrained for four epochs on 8,192 objects. \Cref{tab:ledger} maps every result to its pretrained model, data,
observations and metric.

\subsection{Protocol}
\label{sec:protocol}

\paragraph{Benchmarks.} Three automotive benchmarks fix the inflow condition, so the shape alone determines the field.
WindsorML \citep{ashton2024windsorml} holds 355 variants of
the Windsor body, a simplified car, simulated by wall-modelled large-eddy simulation and split 274/35/35. DrivAerML
\citep{ashton2024drivaerml} supplies car pressure with its official 400/34/50 split and a 420/16/48 split, whose 48 test cars we call reserved. DrivAerNet++ \citep{drivaernetpp} is split 6,492/778/848 by geometry. Two benchmarks leave the
condition unreported. DoubleDelta \citep{doubledelta} holds RANS surface solutions for wings at nine angles of attack,
$11^\circ$--$19^\circ$, at Mach 0.3; its nine-angle split holds out 16 test wings, and two further splits also hold out
angles, between and beyond the training angles. BlendedNet++ \citep{blendednetpp} provides RANS pressure on
blended-wing-body aircraft and tests on unseen geometries at Mach numbers above the training range. \Cref{app:expanded}
gives every split.

\paragraph{Baselines.} Every claim is tested against the same architecture trained from random initialisation, or
\emph{from scratch}, with the same data, observation sets, updates, learning rate and seeds, which isolates what pretraining adds. \Cref{sec:exp-targets} also tunes the learning rate of the model from scratch separately, and the gain
holds. With sensors, the model from scratch is a conditional neural process \citep{garnelo2018cnp}. GABI
\citep{vadeboncoeur2026gabi} learns a prior over fields on varying geometries from simulations and conditions it on the
same sensors by approximate Bayesian computation. It runs its published schedule for three seeds, comparable in
updates (\cref{tab:budgets}), with graph neighbours found along the mesh surface, which suits it better than
straight-line neighbours (\cref{app:impl}). Explicit condition inference evaluates \cref{eq:posterior} with the
true-condition model, which needs condition labels. GeoPT needs the condition as an input, so it is not a baseline for
the sensor task (\cref{app:geopt}).

\paragraph{Metrics and statistics.} Hidden-region nRMSE is the area-weighted RMSE over unobserved points divided by the
case's full-field, area-weighted standard deviation; the shape-only and true-condition references are scored on the
same hidden points as the model they bracket. Force error is the
relative error of the integrated pressure force of the completed field. Means are over three seeds unless
\cref{tab:ledger} lists fewer. Unless stated otherwise, intervals are 95\% paired bootstrap intervals over physical
geometries, Bonferroni-adjusted where one study makes several comparisons. Hidden-value perturbation checks guard
against leakage, and scores are recomputed independently from saved predictions (\cref{app:controls}).

\subsection{Half the simulations to learn the same field}
\label{sec:exp-pretrain}

We count the training simulations each model needs to reach a test nRMSE of 0.25 without sensors, trained on a
sixteenth to all of each training set. \areals{} needs 35 WindsorML, 200 DrivAerML and 1,623 DrivAerNet++
simulations, against 137, 400 and 3,246 from scratch: 74.5\%, 50\% and 50\% fewer (\cref{fig:efficiency}). It has the
lower error at every budget, and needs fewer simulations in every bootstrap draw on WindsorML and DrivAerML and in
99.9\% on DrivAerNet++. The gain is largest where simulations are scarcest and narrows as they grow, as a prior's
should (\cref{app:expanded}).

The savings are similar at a target of 0.3 and with 20 sensors; at 0.2, only \areals{} reaches the target on
DrivAerML (\cref{app:expanded}). Nor do they need the full corpus: the 1,668-mesh prior of \cref{sec:exp-targets} also
reaches 0.25 on WindsorML with 35 cars. Counting validation simulations as well, the savings are 40--59\%. Budgets are the smallest tested at 8,000 updates per fit. On DrivAerML that is four times the 40-epoch schedule of
\cref{tab:pretrain}, and the gap stays open, 0.181 against 0.219 with 400 cars. On DrivAerNet++ it is a quarter of that
schedule, and 17 of 30 fits were still improving, so converged training could change the savings there.

The prior keeps paying when simulations are plentiful. With the 40-epoch schedule it lowers the error by 18.0\% with
all 400 official DrivAerML cars, by 20.2\% on the 420-car split and by 2\% with all 6,492 DrivAerNet++ cars, every
paired interval excluding zero (\cref{tab:pretrain,tab:expanded}). As frozen inputs, areal tokens also lower the error
of a surrogate they were not trained with, by 0.9\% (\cref{app:frozen}).

\subsection{Distance and normal line carry the transfer}
\label{sec:exp-targets}

\Cref{sec:el} predicts which targets transfer. Distance and normal line, which fix the flow equation's coefficients to first
order, should transfer.
Curvature, which enters only through their derivatives, should add little. The coefficients alone should not suffice, since they live in wall coordinates that only the global distance field supplies. Three
priors test this, pretrained from fresh weights without masking on the same 1,668 Objaverse-XL meshes for 10,626 updates (\cref{app:targets}). They are \areals{} ($d,\nnT$); an \areals{}-sized network that adds the off-surface Hessian of $\eta_S$,
which carries the shape operator; and a patch encoder trained on the coefficients $A$ of smooth local charts, with no
global distance field. Each is fine-tuned on WindsorML pressure from geometry alone against its own architecture from
scratch, with the learning rate tuned for every arm, initialisation and budget.

\begin{table}[t]
\caption{Matched pretraining targets on WindsorML: test nRMSE ($\downarrow$) without sensors, pretrained / from scratch,
means of three seeds over the 35 test cars. Each column is a prior pretrained on the same 1,668 meshes, against its own
architecture from scratch (\cref{sec:exp-targets}); settings were fixed on validation before testing, and intervals
are in \cref{tab:targets-full}.}
\label{tab:targets}
\centering
\footnotesize
\setlength{\tabcolsep}{3pt}
\begin{tabular}{@{}rccc@{}}
\toprule
Cars & $d,\nnT$ (\areals{}) & $d,\nnT$, Hessian & Patch $A$ \\
\midrule
18 & \textbf{0.277} / 0.443 & 0.301 / 0.475 & 0.457 / 0.442 \\
35 & \textbf{0.228} / 0.381 & 0.246 / 0.388 & 0.384 / 0.379 \\
69 & \textbf{0.193} / 0.265 & 0.206 / 0.301 & 0.307 / 0.273 \\
137 & \textbf{0.178} / 0.228 & 0.179 / 0.215 & 0.257 / 0.223 \\
274 & \textbf{0.158} / 0.203 & 0.161 / 0.182 & 0.221 / 0.196 \\
\bottomrule
\end{tabular}
\end{table}

All three predictions hold on the 35 test cars (\cref{tab:targets}). Distance and normal line cut the error by
22--40\% at every budget, on at least 31 of the 35 cars. Adding the Hessian buys nothing: that network is never more
accurate than \areals{} and is less accurate at the three smallest budgets, though as a separate implementation its extra
target also costs accuracy in distance and normals (\cref{tab:target-recipes}). The patch
encoder reproduces the coefficients to 1.9\% and normals to $1.6^\circ$, yet transfers worse than scratch at every
budget, significantly at the three largest; it needs 274 cars to reach 0.25, where scratch needs 137 and \areals{} 35.

\paragraph{The global distance field decides transfer.} The patch encoder differs from \areals{} in architecture as
well as target. A fourth prior removes the difference: it trains the same coefficients inside \areals{}'s own encoder
and query decoder, beside distance and normal line, on 14,304 meshes. It transfers as \areals{} does, 18--38\% below
scratch at every budget and within 0.007 nRMSE of the same network trained without the coefficients
(\cref{app:targets}). Accurate local coefficients are not enough; what transfers is the global distance field from
which wall coordinates and coefficients are both read.

\paragraph{Reconstruction, not masking, is what transfers.} As \cref{prop:sufficiency} predicts, the same network
pretrained on whole surfaces transfers as well as with masking: every interval over three pretraining seeds and all
sensor counts contains zero (\cref{tab:mask}). The matched priors above are pretrained this way.

\subsection{Sensors in place of the operating condition}
\label{sec:exp-sensors}
\label{sec:exp-dd}
\label{sec:exp-posterior}

\begin{table}[t]
\caption{Sensor-conditioned comparison at 20 sensors with the condition unreported: hidden-region nRMSE ($\downarrow$)
on unseen geometries. Scratch is the \areals{} sensor model without geometry pretraining; the shape-only (Shape) and
true-condition (Condition) references see no sensors. GABI was run on the nine-angle split only. Means of three seeds,
except one seed for the nine-angle model from scratch and the references.}
\label{tab:sensors}
\centering
\footnotesize
\setlength{\tabcolsep}{2.5pt}
\begin{tabular}{@{}lccccc@{}}
\toprule
 & \multicolumn{3}{c}{Sensor-conditioned} & \multicolumn{2}{c}{References} \\
\cmidrule(lr){2-4}\cmidrule(l){5-6}
Benchmark & GABI & Scratch & \areals{} & Shape & Condition \\
\midrule
DoubleDelta, nine angles & 0.642 & 0.222 & \textbf{0.164} & 0.302 & 0.137 \\
DoubleDelta, $18$--$19^\circ$ & --- & 0.409 & \textbf{0.206} & 0.373 & 0.193 \\
BlendedNet++ & --- & 0.212 & \textbf{0.176} & 0.743 & 0.066 \\
\bottomrule
\end{tabular}
\end{table}

On the nine-angle DoubleDelta split at 20 sensors, \areals{} has 3.9 times lower error than GABI, 4.0 times lower integrated-force error,
and 26.2\% lower error than the model from scratch (\cref{tab:sensors}). Against GABI, the error ratio stays at
3.5--3.9 from 5 to 500 sensors, every GABI seed is worse than every \areals{} seed at every setting, and GABI already
trails with nothing observed (\cref{tab:gabi}).

\paragraph{Sensors recover most of what the condition is worth.} On the nine-angle split, 20 sensors close 84\% of the
gap between the shape-only and true-condition models in nRMSE (\cref{sec:levels}), and \gapClosed{} in normalised MSE for
the first fine-tuning seed (95\% CI \gapCI{}). At 100 sensors \areals{} comes within 0.005 nRMSE of the true-condition
model, and at 500 the two are level (\cref{fig:doubledelta}). The integrated force follows: its error falls from 15.5\%
for the shape-only model to 3.7\% at 20 sensors and 1.5\% at 500, below the true-condition model's 1.8\%. Where there is
no condition to recover, as on DrivAerML at its one inflow, $\E[f\mid S,y]=\E[f\mid S]$, and the sensor model matches
the shape-only model to within 0.002 nRMSE at every sensor count (\cref{tab:expanded}).

\paragraph{Beyond the training range.}\label{sec:exp-beyond} On unseen wings at $18$--$19^\circ$, beyond the training
angles, 20 sensors close 93\% of the gap in nRMSE, and from 100 sensors the error of \areals{} is below that of the
true-condition model (\cref{tab:sensors,tab:expanded}). The model from scratch cannot use its sensors there: its error
stays at 0.41--0.42 at every sensor count, above the shape-only model's, so pretraining halves the error at 20 sensors
and cuts it by 57\% at 100. The failure is specific to extrapolation, since on the nine-angle split the same model reads
its sensors (\cref{tab:sensors}). Explicit inference over the training angles cannot follow the angle either: it stays
at 0.300--0.307 from 20 to 500 sensors (\cref{tab:explicit}), and a wider search could at best reach the
true-condition model.

\paragraph{Other unseen conditions.} On BlendedNet++, where the flight condition dominates the field, 20 sensors close
84\% of the gap in nRMSE on unseen aircraft at higher Mach numbers. Pretraining lowers the error of the model from scratch there by 15--27\% from 5 to 500 sensors (\cref{tab:expanded}). Between the training angles, 20 sensors close
\heldGap{} of the gap in normalised MSE (\cref{tab:held-angle}).

\paragraph{Where the posterior is exact.} On Gaussian fields drawn on 16 held-out shape families, whose posterior mean
is exact, fine-tuned \areals{} recovers 82.6\% of that mean's gain over predicting with nothing observed, against 25.4\% from scratch. With its geometry trunk frozen at $\theta_{\mathrm{geo}}$ it still recovers 30.4\%, six times the 4.7\%
of a frozen random trunk (\cref{tab:posterior}).

\subsection{Transfer runs through local attention and the query decoder}
\label{sec:exp-mechanism}
\label{sec:exp-decoder}

\paragraph{Pretraining makes attention local; simulations extend its reach.} Decoder attention on the 16 DoubleDelta
test wings bears out the division of work of \cref{sec:levels} (\cref{fig:attention,tab:attention}). The pretrained
decoder places 52\% of its attention within $0.25R$ of the query, with $R$ the surface's RMS radius, where attention in
proportion to area would place 3\%. Fine-tuning raises its share beyond $0.5R$ from 17\% to 40\% on every wing, while
the model trained from scratch spreads its attention almost in proportion to area, with 72\% beyond $0.5R$. The gain
is not confined to curved regions: \areals{} removes a similar share of the reducible error of the model from scratch on the most curved and the flattest
thirds of every wing (\cref{app:attention}).

\paragraph{The transfer runs through the query decoder.} With fresh query decoders, a released Point-MAE encoder nearly
ten times the size of an \areals{} encoder pretrained for four epochs has the lower error on 40 DrivAerML cars withheld
from training, 0.256 against 0.271. Keeping the pretrained query decoder reverses this: 0.240, \decoderGain{} below
resetting it in all three seeds and \pointMaeGain{} below Point-MAE, both adjusted intervals excluding zero
(\cref{tab:decoder}). Pretraining teaches the decoder how to read the surface, not only the encoder what it contains.

\subsection{Each design choice does its job}
\label{sec:exp-design}
\label{sec:exp-quadrature}
\label{sec:exp-geometry}

\paragraph{The distance target reproduces topology.} As \cref{lem:primitive} predicts, the distance to the token disks
reproduces every component, loop and cavity that persists beyond 0.21 RMS radii on all eleven test shapes. With 60\% of the area hidden, the learned distance stays below 0.05 RMS radii over 79--98\% of each held-out mesh's hidden
area, against 8--16\% for disks built from the visible surface (\cref{app:tubes}).

\paragraph{The sign-free target recovers normal lines best.} On 128 held-out objects over three seeds, the projector
target reaches an unoriented angle error of $25.3^\circ$, against $28.7^\circ$ for an oriented-vector target and
$39.2^\circ$ for an unoriented-vector target (\cref{app:architecture}).

\paragraph{Area weights make attention nearly blind to sampling.} One half of each of 32 meshes is sampled nine times more densely, with weights that
represent the same area. Log-area attention then changes the normalised output by 0.0025,
against 0.0137 for ordinary attention, \densityReduction{} less drift; the 95\% interval for the difference excludes
zero (\cref{app:stability}).

\paragraph{Surface-anchored, rotary-positioned tokens cut error by over a third.} Anchoring the tokens to the surface
together with rotary positions lowers full-surface error on the 48 reserved cars from 0.390 to 0.247, a
36.7\% reduction with adjusted interval 35.0--38.1\% (\cref{app:architecture}).

\section{Related Work}
\label{sec:related}

\paragraph{Masked 3D pretraining and shape autoencoders.} Masked point modelling predicts hidden patches or classifies
query points \citep{yu2022pointbert,pang2022pointmae,liu2022maskpoint,boulch2023also}, and MaskSurf and MaskFeat3D
predict surfels and normals \citep{zhang2022masksurf,yan2024maskfeat3d}; all treat completion as a pretext task and
discard the decoder. Shape autoencoders decode latent sets by cross-attention
\citep{zhang2023shape2vecset,zhang20223dilg,lattice2026}, but occupancy and signed-distance targets need watertight
meshes \citep{zhang2024clay,chen2025dora}, which unsigned distance avoids \citep{chibane2020ndf,yan2023iae,log3d2026};
GeoUDF and 2D Gaussian splatting fit tangent planes or disks per shape \citep{ren2023geoudf,huang20242dgs}. \areals{}
treats completion as the task itself, queries sign-free targets anywhere in space, reads its tokens as surfels, and
transfers its query decoder to physical fields.

\paragraph{Surrogates, reconstruction and inverse problems.} Simulation surrogates learn geometry inside a supervised
model \citep{li2023gino,wu2024transolver,alkin2025abupt}, or pretrain on large simulation datasets and adapt to new
configurations \citep{zou2026adafield,yang2026aerotransformer}. Others pretrain without simulation labels
\citep{geoca2024,chen2024pretraining,zhou2024maepde}: \citet{zhang2026frozen} pretrain a shape autoencoder on a
geometry-reconstruction target chosen without reference to the physics, and GeoPT adds synthetic dynamics
\citep{geopt2026}. None of these takes measurements. Gappy proper orthogonal decomposition, the parametrised-background
data-weak method, Senseiver, BLISSNet and sequential estimators reconstruct flows from sparse sensors on fixed or
two-dimensional domains \citep{everson1995gappy,maday2015pbdw,santos2023senseiver,blissnet2026,seqest2025}, and GABI
conditions a learned field prior on observations by approximate Bayesian computation \citep{vadeboncoeur2026gabi}.
Unlike all of these, \areals{}
reads sparse sensors with a model pretrained on raw, unlabelled meshes to reconstruct targets derived from the
governing equation.

\paragraph{Measure-aware attention and symmetry.} Log-weights in attention logits appear as proportional attention in
ToMe \citep{bolya2023tome}; attention can be read as a kernel integral \citep{kovachki2023neural,cao2021galerkin} whose
continuum limit assumes uniform sampling \citep{calvello2025continuum}, and quadrature weights enter neural operators
and point convolutions outside attention \citep{pcno2025,quadnorm2026,hermosilla2018mcconv,sharp2022diffusionnet}.
Symmetry-aware completion copies or fuses visible geometry through detected or learned symmetries
\citep{mitra2006symmetry,li2023e3sym,thrun2005shape,mavridis2015completion,ladybird2020,symmcompletion2025,patchrd2022,zhao2026symmetry}.
\areals{} weights every attention layer by surface area and adds each voted rigid symmetry as an observation operator.


\section{Conclusion}
\label{sec:conclusion}

Euler--Lagrange geometry pretraining derives its targets, distance and normal line, from the flow equation and learns
them from raw meshes alone. Compared with the same architecture trained from scratch, \areals{} learns the same field
from 50--74.5\% fewer simulations. In a matched comparison of targets, distance and normal line carry the transfer:
accurate local coefficients add nothing to them and, on their own, transfer worse than scratch. Geometry supplies the field's local information and simulations assemble the global field: pretrained
attention is local, and fine-tuning extends its reach. Sensors then supply the operating condition; on unseen wings, 20
sensors recover 84\% of what knowing the true angle is worth, at 3.9 times lower error than GABI.

\paragraph{Limitations.} The derivation is exact only for attached, irrotational flow; for the separated, turbulent
flows we test, the evidence for the targets is empirical. The savings are measured at a fixed training budget, and
every benchmark is external aerodynamics.

Abundant shapes pay for scarce simulations.

\section*{Impact Statement}
This work reduces the simulation and testing that engineering design requires by inferring surface fields from
geometry and sparse measurements, lowering the computational and energy cost of each design iteration. As with any
learned estimate, completed fields used in safety-critical decisions should be confirmed by simulation or measurement.

\bibliography{refs,refs_extra}
\IfFileExists{icml2026.bst}{\bibliographystyle{icml2026}}{\bibliographystyle{plainnat}}

\newpage
\appendix
\onecolumn
\counterwithin{figure}{section}
\counterwithin{table}{section}

\section{Experiment Map}
\label{app:ledger}

\begin{table}[H]
\caption{Where every result comes from. ``Production'' is the \nMeshes{}-mesh geometry checkpoint selected at epoch 11;
``controlled'' denotes the four-epoch checkpoints pretrained on 8,192 objects (\cref{app:impl}). Observations list what
the model sees at test time; hidden nRMSE scores only unobserved points.}
\label{tab:ledger}
\centering
\scriptsize
\setlength{\tabcolsep}{3pt}
\begin{tabular}{@{}>{\raggedright\arraybackslash}p{3.0cm}>{\raggedright\arraybackslash}p{2.8cm}>{\raggedright\arraybackslash}p{2.3cm}>{\raggedright\arraybackslash}p{2.6cm}>{\raggedright\arraybackslash}p{2.0cm}>{\raggedright\arraybackslash}p{1.9cm}>{\raggedright\arraybackslash}p{1.2cm}@{}}
\toprule
Result & Geometry checkpoint & Field training & Evaluation & Observations & Metric & Seeds \\
\midrule
Unreported angle (\cref{fig:doubledelta}, \cref{tab:sensors,tab:gabi,tab:dd-full,tab:gabi-full}) & Production vs.\ none & 2,160 cases, 40 epochs per model & 144 test cases (16 wings) & 0--500 sensors; patches & Hidden nRMSE; force error & 3; scratch and references 1 \\
Held-out angles (\cref{tab:held-angle}) & Production & 1,200 cases (odd angles) & 64 cases (16 wings, even angles) & 5--500 sensors & Hidden nRMSE & 1 \\
Explicit condition inference (\cref{tab:explicit}) & Production & True-condition model, 228 wing families, $11$--$16^\circ$ & 16 wings at $18$--$19^\circ$; timing on 4 validation wings & 20--500 sensors & Hidden nRMSE; time per case & 3 \\
Extrapolated angles (\cref{tab:sensors,tab:pretrain,tab:expanded}) & Production vs.\ none & 228 wing families, $11$--$16^\circ$ & 16 wings at $18$--$19^\circ$ & 5--500 sensors & Hidden nRMSE & 3 \\
BlendedNet++ (\cref{tab:sensors,tab:pretrain,tab:expanded}) & Production vs.\ none & 8,736 aircraft & 1,872 unseen aircraft, higher Mach & 5--500 sensors & Hidden nRMSE & 3 \\
DrivAerML official split (\cref{tab:pretrain,tab:expanded}) & Production vs.\ none & 400 cars & 50 test cars & 5--500 sensors & Hidden nRMSE & 3 \\
DrivAerNet++ (\cref{tab:pretrain,tab:expanded}) & Production vs.\ none & 6,492 cars & 848 test cars & 5--500 sensors & Hidden nRMSE & 3 \\
Simulation efficiency (\cref{fig:efficiency}) & Production vs.\ none & A sixteenth to all of 274 WindsorML, 400 DrivAerML and 6,492 DrivAerNet++ simulations; 8,000 updates per fit & 35, 50 and 848 test cases & None; 20 sensors & Test nRMSE; budget reaching 0.25 & 3 \\
Masked vs.\ unmasked (\cref{tab:mask}) & Masked and unmasked, seeds 17--19, epoch 11 & 228 wing families; 400 cars & 16 wings at $18$--$19^\circ$; 50 test cars & 0--500 sensors & Hidden nRMSE & 3 \\
Reserved-car completion (\cref{tab:pretrain}) & Production vs.\ none & 420 cars & 48 reserved cars & None; 25 settings & Hidden nRMSE & 3 \\
Gaussian fields (\cref{tab:posterior,tab:gaussian-full}) & \areals{} pretraining vs.\ random trunk & Gaussian fields, 9,120 updates & 16 held-out families, 8 fields each & 5--500 sensors; patches & Share of posterior gain & 3 \\
Decoder transfer and head initialisation (\cref{tab:decoder}, \cref{app:architecture}) & Controlled; released Point-MAE & 380 cars & 40 cars withheld from the training pool & None & Full-surface nRMSE & 3 \\
Coordinate interface (\cref{app:architecture}) & Controlled variants & Car training pool & 48 reserved cars & None & Full-surface nRMSE & 3 \\
Normal targets (\cref{app:architecture}) & Controlled variants & --- & 128 held-out objects & Geometry & Normal angle & 3 \\
Density shift (\cref{sec:exp-quadrature}, \cref{app:stability}) & Three pretraining seeds per attention type & --- & 32 meshes & Geometry & Output drift; distance NMSE & 3 \\
Finite tokens (\cref{app:stability}) & Production & --- & 32 meshes & Geometry & Output change & 1 \\
Symmetry (\cref{tab:geometry}) & Surfel readouts; gate trained on 170 meshes & --- & 41 held-out meshes & Full; 60\% hidden & Normal error & 1 \\
Distance tubes (\cref{tab:tubes,tab:tubes-masked}) & Production & --- & 8 held-out meshes; sphere, torus, two spheres & Full; 60\% hidden & Tube area, Betti numbers, paths, bottleneck; coverage & 1 \\
Attention range and field error by region (\cref{fig:attention}, \cref{tab:attention}) & Production vs.\ none & 2,160 cases, 40 epochs per model & 144 test cases (16 wings) & None & Attention distance and mass; error reduction & 1 \\
Frozen surrogate (\cref{app:frozen}) & Earlier \nMeshes{}-mesh checkpoint (epoch 14) & AeroJEPA tutorial, 420 cars & 16 validation cars; 48 reserved cars & None & Relative $L_2$ & 3 on 48 cars \\
Matched pretraining targets (\cref{tab:targets,tab:targets-full,tab:target-recipes}) & Three fresh priors ($d,\nnT$; $d,\nnT,H_r$; patch $A$) on 1,668 meshes, 10,626 updates & WindsorML, 18--274 cars, 8,000 updates per fit & 35 test cars, locked after validation selection; 92 test meshes for geometry & None & Hidden nRMSE; geometry errors & 3 \\
Patch pilot (\cref{app:targets}) & Patch prior on 385 meshes & 274 cars & 35 test cars & None & Hidden nRMSE & 1 \\
Coefficients inside \areals{} (\cref{tab:coefficient-prior}) & \areals{} priors on 14,304 meshes, with $A$ (seeds 17, 29, 43) and without (17, 29) & WindsorML, 18--274 cars, 8,000 updates per fit, rate per seed & 35 test cars & None & Hidden nRMSE & 3; 1 per pretraining seed \\
\bottomrule
\end{tabular}
\end{table}

\section{Formal Details}
\label{app:formal}

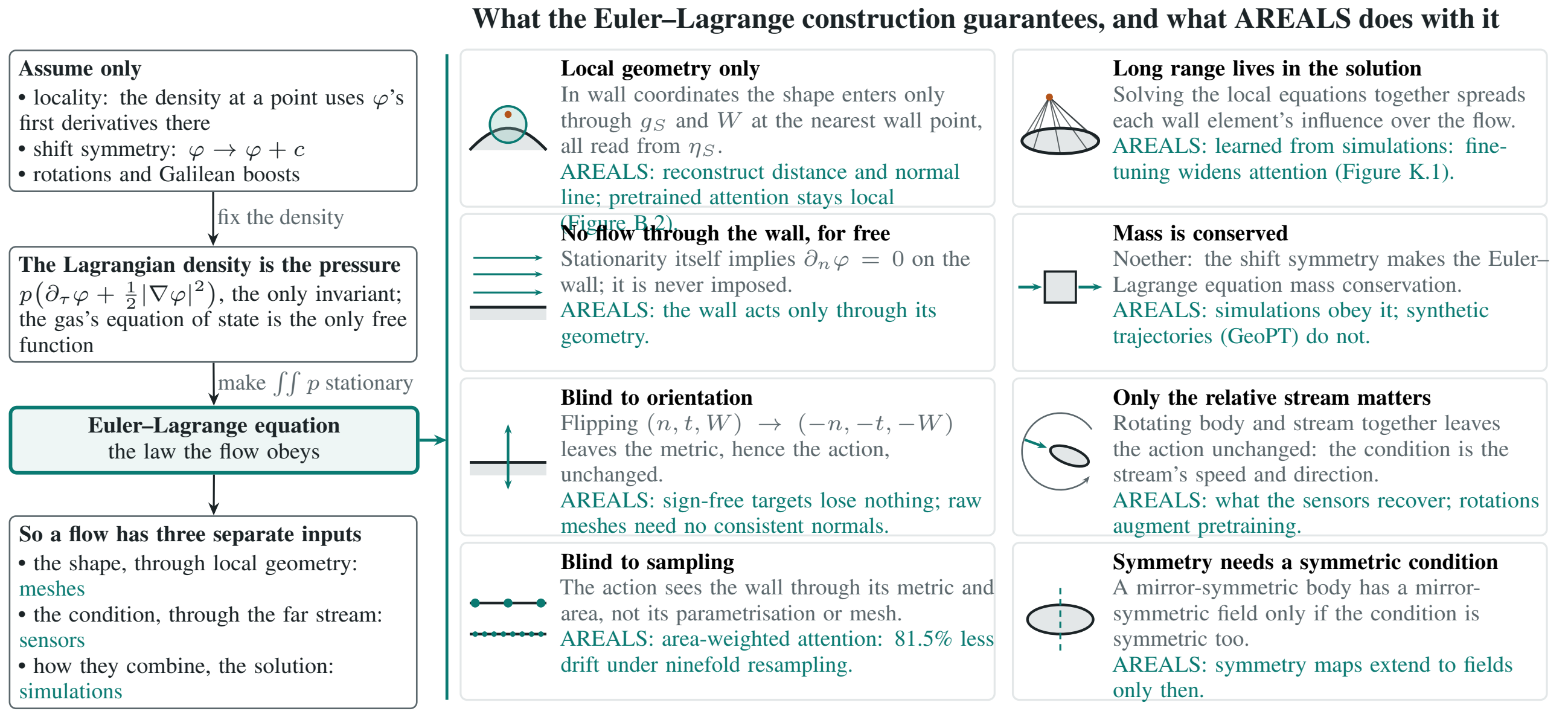
\begin{figure}[H]
\centering
\begin{tikzpicture}[x=1cm, y=1cm, font=\scriptsize, execute at begin node={\hyphenpenalty=10000},
    box/.style={draw=arealmuted, line width=0.5pt, rounded corners=2pt, fill=white, inner sep=3pt, align=left,
                text width=4.25cm, font=\scriptsize, text=arealfg},
    elbox/.style={box, draw=arealteal, line width=1.1pt, fill=arealteal!7, align=center},
    flow/.style={-{Stealth[length=4pt,width=3.4pt]}, line width=0.6pt, draw=arealfg},
    lab/.style={inner sep=1pt, font=\scriptsize, text=arealmuted}]
\node[box, anchor=north west] (A) at (0.05,6.85)
  {\textbf{Assume only}\\[1pt]
   \textbullet\ locality: the density at a point uses $\varphi$'s first derivatives there\\
   \textbullet\ shift symmetry: $\varphi\to\varphi+c$\\
   \textbullet\ rotations and Galilean boosts};
\node[box, anchor=north west] (B) at (0.05,4.7)
  {\textbf{The Lagrangian density is the pressure}\\[1pt]
   $p\big(\partial_\tau\varphi+\tfrac12|\nabla\varphi|^2\big)$, the only invariant; the gas's equation of state is the
   only free function};
\node[elbox, anchor=north west] (C) at (0.05,2.95)
  {\textbf{Euler--Lagrange equation}\\the law the flow obeys};
\node[box, anchor=north west] (D) at (0.05,1.75)
  {\textbf{So a flow has three separate inputs}\\[1pt]
   \textbullet\ the shape, through local geometry: \textcolor{arealteal}{meshes}\\
   \textbullet\ the condition, through the far stream: \textcolor{arealteal}{sensors}\\
   \textbullet\ how they combine, the solution: \textcolor{arealteal}{simulations}};
\draw[flow] (A) -- (B) node[lab, midway, right] {fix the density};
\draw[flow] (B) -- (C) node[lab, midway, right] {make $\iint p$ stationary};
\draw[flow] (C) -- (D);
\node[anchor=south west, font=\small\bfseries, text=arealfg] at (5.0,6.9) {What the Euler--Lagrange construction guarantees, and what \areals{} does with it};
\draw[flow, line width=0.9pt, draw=arealteal] (C.east) -- (C.east -| 4.85,0);
\draw[arealteal, line width=0.9pt] (4.85,-0.27) -- (4.85,6.8);
\draw[arealgrid, line width=0.6pt, rounded corners=2pt] (5.00,6.85) rectangle (10.85,5.13);
\fill[arealgrid] (5.10,5.75) .. controls (5.40,6.07) and (5.64,6.07) .. (5.94,5.75) -- cycle;
\draw[arealfg, line width=0.9pt] (5.10,5.75) .. controls (5.40,6.07) and (5.64,6.07) .. (5.94,5.75);
\draw[arealteal, line width=0.7pt, fill=arealrec, fill opacity=0.3] (5.52,6.03) circle (0.2);
\fill[arealwarn] (5.52,6.14) circle (1.1pt);
\node[anchor=north west, align=left, text width=4.75cm, inner sep=2pt] at (6.02,6.81)
  {\textbf{Local geometry only}\\\textcolor{arealmuted}{In wall coordinates the shape enters only through $g_S$ and $W$ at the nearest wall point, all read from $\eta_S$.}\\\textcolor{arealteal}{\areals{}: reconstruct distance and normal line; pretrained attention stays local (\cref{fig:locality}).}};
\draw[arealgrid, line width=0.6pt, rounded corners=2pt] (5.00,5.05) rectangle (10.85,3.33);
\fill[arealgrid] (5.10,3.89) rectangle (5.94,4.03);
\draw[arealfg, line width=0.9pt] (5.10,4.03) -- (5.94,4.03);
\draw[arealteal, line width=0.6pt, -{Stealth[length=2.5pt]}] (5.14,4.19) -- (5.90,4.19);
\draw[arealteal, line width=0.6pt, -{Stealth[length=2.5pt]}] (5.14,4.39) -- (5.90,4.39);
\draw[arealteal, line width=0.6pt, -{Stealth[length=2.5pt]}] (5.14,4.57) -- (5.90,4.57);
\node[anchor=north west, align=left, text width=4.75cm, inner sep=2pt] at (6.02,5.01)
  {\textbf{No flow through the wall, for free}\\\textcolor{arealmuted}{Stationarity itself implies $\partial_n\varphi=0$ on the wall; it is never imposed.}\\\textcolor{arealteal}{\areals{}: the wall acts only through its geometry.}};
\draw[arealgrid, line width=0.6pt, rounded corners=2pt] (5.00,3.25) rectangle (10.85,1.53);
\fill[arealgrid] (5.10,2.19) rectangle (5.94,2.33);
\draw[arealfg, line width=0.9pt] (5.10,2.33) -- (5.94,2.33);
\draw[arealteal, line width=0.8pt, {Stealth[length=3pt]}-{Stealth[length=3pt]}] (5.52,2.03) -- (5.52,2.75);
\node[anchor=north west, align=left, text width=4.75cm, inner sep=2pt] at (6.02,3.21)
  {\textbf{Blind to orientation}\\\textcolor{arealmuted}{Flipping $(n,t,W)\to(-n,-t,-W)$ leaves the metric, hence the action, unchanged.}\\\textcolor{arealteal}{\areals{}: sign-free targets lose nothing; raw meshes need no consistent normals.}};
\draw[arealgrid, line width=0.6pt, rounded corners=2pt] (5.00,1.45) rectangle (10.85,-0.27);
\draw[arealfg, line width=0.8pt] (5.10,0.79) -- (5.94,0.79);
\fill[arealteal] (5.160,0.79) circle (1.4pt);
\fill[arealteal] (5.520,0.79) circle (1.4pt);
\fill[arealteal] (5.880,0.79) circle (1.4pt);
\draw[arealfg, line width=0.8pt] (5.10,0.45) -- (5.94,0.45);
\fill[arealteal] (5.160,0.45) circle (0.85pt);
\fill[arealteal] (5.263,0.45) circle (0.85pt);
\fill[arealteal] (5.366,0.45) circle (0.85pt);
\fill[arealteal] (5.469,0.45) circle (0.85pt);
\fill[arealteal] (5.571,0.45) circle (0.85pt);
\fill[arealteal] (5.674,0.45) circle (0.85pt);
\fill[arealteal] (5.777,0.45) circle (0.85pt);
\fill[arealteal] (5.880,0.45) circle (0.85pt);
\node[anchor=north west, align=left, text width=4.75cm, inner sep=2pt] at (6.02,1.41)
  {\textbf{Blind to sampling}\\\textcolor{arealmuted}{The action sees the wall through its metric and area, not its parametrisation or mesh.}\\\textcolor{arealteal}{\areals{}: area-weighted attention: 81.5\% less drift under ninefold resampling.}};
\draw[arealgrid, line width=0.6pt, rounded corners=2pt] (11.05,6.85) rectangle (16.90,5.13);
\fill[arealgrid] (11.57,5.85) ellipse (0.42 and 0.14);
\draw[arealfg, line width=0.35pt, opacity=0.7] (11.45,6.33) -- (11.990,5.850);
\draw[arealfg, line width=0.35pt, opacity=0.7] (11.45,6.33) -- (11.867,5.949);
\draw[arealfg, line width=0.35pt, opacity=0.7] (11.45,6.33) -- (11.570,5.990);
\draw[arealfg, line width=0.35pt, opacity=0.7] (11.45,6.33) -- (11.273,5.949);
\draw[arealfg, line width=0.35pt, opacity=0.7] (11.45,6.33) -- (11.150,5.850);
\draw[arealfg, line width=0.35pt, opacity=0.7] (11.45,6.33) -- (11.273,5.751);
\draw[arealfg, line width=0.35pt, opacity=0.7] (11.45,6.33) -- (11.570,5.710);
\draw[arealfg, line width=0.35pt, opacity=0.7] (11.45,6.33) -- (11.867,5.751);
\draw[arealfg, line width=0.9pt] (11.57,5.85) ellipse (0.42 and 0.14);
\fill[arealwarn] (11.45,6.33) circle (1.1pt);
\node[anchor=north west, align=left, text width=4.75cm, inner sep=2pt] at (12.07,6.81)
  {\textbf{Long range lives in the solution}\\\textcolor{arealmuted}{Solving the local equations together spreads each wall element's influence over the flow.}\\\textcolor{arealteal}{\areals{}: learned from simulations: fine-tuning widens attention (\cref{fig:attention}).}};
\draw[arealgrid, line width=0.6pt, rounded corners=2pt] (11.05,5.05) rectangle (16.90,3.33);
\draw[arealfg, line width=0.7pt, fill=arealgrid] (11.40,4.08) rectangle (11.74,4.42);
\draw[arealteal, line width=0.9pt, -{Stealth[length=3pt]}] (11.11,4.25) -- (11.37,4.25);
\draw[arealteal, line width=0.9pt, -{Stealth[length=3pt]}] (11.77,4.25) -- (12.03,4.25);
\node[anchor=north west, align=left, text width=4.75cm, inner sep=2pt] at (12.07,5.01)
  {\textbf{Mass is conserved}\\\textcolor{arealmuted}{Noether: the shift symmetry makes the Euler--Lagrange equation mass conservation.}\\\textcolor{arealteal}{\areals{}: simulations obey it; synthetic trajectories (GeoPT) do not.}};
\draw[arealgrid, line width=0.6pt, rounded corners=2pt] (11.05,3.25) rectangle (16.90,1.53);
\begin{scope}[rotate around={-20:(11.67,2.40)}]
\draw[arealfg, line width=0.8pt, fill=arealgrid] (11.67,2.40) ellipse (0.22 and 0.09);
\draw[arealteal, line width=0.8pt, -{Stealth[length=3pt]}] (11.15,2.40) -- (11.41,2.40);
\end{scope}
\draw[arealmuted, line width=0.5pt, -{Stealth[length=2.5pt]}] (11.57,2.45) ++(35:0.42) arc (35:320:0.42);
\node[anchor=north west, align=left, text width=4.75cm, inner sep=2pt] at (12.07,3.21)
  {\textbf{Only the relative stream matters}\\\textcolor{arealmuted}{Rotating body and stream together leaves the action unchanged: the condition is the stream's speed and direction.}\\\textcolor{arealteal}{\areals{}: what the sensors recover; rotations augment pretraining.}};
\draw[arealgrid, line width=0.6pt, rounded corners=2pt] (11.05,1.45) rectangle (16.90,-0.27);
\draw[arealfg, line width=0.8pt, fill=arealgrid] (11.57,0.61) ellipse (0.36 and 0.16);
\draw[arealteal, line width=0.6pt, dash pattern=on 2pt off 1.5pt] (11.57,0.27) -- (11.57,0.97);
\node[anchor=north west, align=left, text width=4.75cm, inner sep=2pt] at (12.07,1.41)
  {\textbf{Symmetry needs a symmetric condition}\\\textcolor{arealmuted}{A mirror-symmetric body has a mirror-symmetric field only if the condition is symmetric too.}\\\textcolor{arealteal}{\areals{}: symmetry maps extend to fields only then.}};
\end{tikzpicture}
\caption{\textbf{The Euler--Lagrange construction and its guarantees.} Assuming only locality, the shift symmetry,
rotations and Galilean boosts fixes the flow's Lagrangian density as the pressure, with the gas's equation of state the
only free function; making its integral stationary gives the Euler--Lagrange equation, the law the flow obeys. Each
guarantee on the right follows from this construction, and its last line says what \areals{} builds in or learns from
it.}
\label{fig:guarantees}
\end{figure}

\begin{figure}[H]
\centering
\input{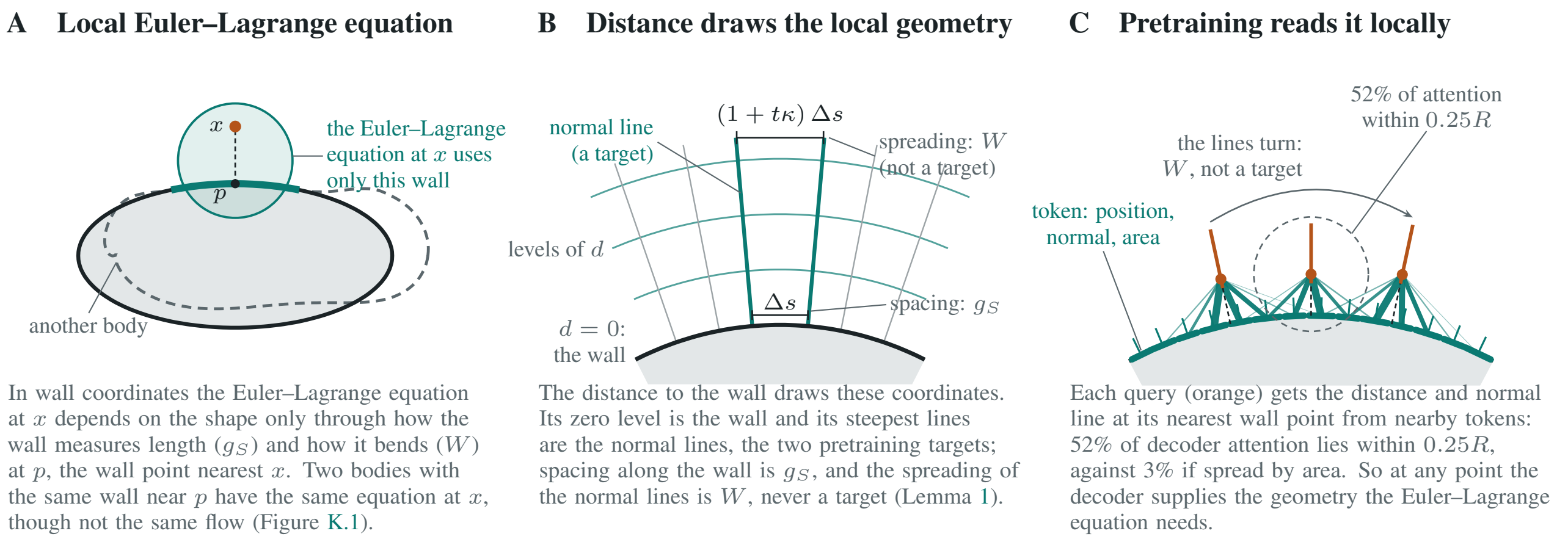}
\caption{\textbf{Near the wall the Euler--Lagrange equation needs only local geometry, the distance encodes it, and
pretraining reads it} (schematic; computed counterparts in \cref{fig:hero}). (A) Bodies with the same wall around $p$,
the wall point nearest $x$, have the same Euler--Lagrange equation at $x$, though not the same flow: in wall coordinates
the Lagrangian density, the pressure times $\sqrt g$, sees the shape only through the surface metric $g_S$ and the shape
operator $W$ at $p$. (B) The level sets and steepest lines of the distance are those coordinates; the value, Hessian and
third derivative of $\eta_S=\tfrac12 d_S^2$ give the wall, its normal lines and their spreading $W$
(\cref{lem:primitive}). (C) The pretrained decoder reads the distance and normal line at each query's nearest wall
point from nearby tokens; $W$ is not supervised. Widths and the dashed radius in C are schematic; the 52\% is measured
on 16 DoubleDelta test wings (\cref{app:attention}).}
\label{fig:locality}
\end{figure}

\begin{figure}[H]
\centering
\includegraphics[width=\textwidth]{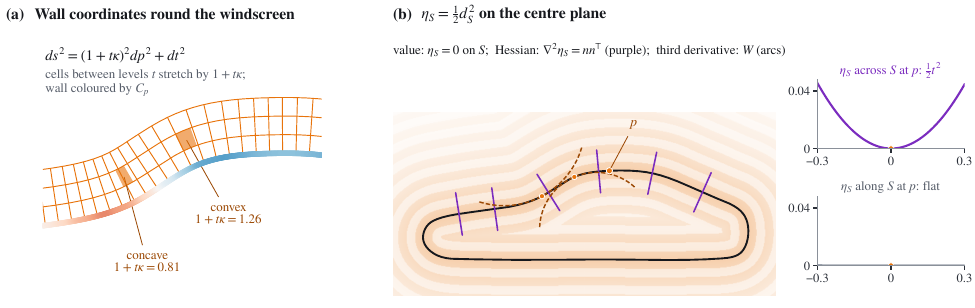}
\caption{\textbf{Wall coordinates and the half squared distance on the car of \cref{fig:hero}.} (a)~Round the
windscreen, cells between levels of $t$ stretch by $1+t\kappa$: to 0.81 over the concave and 1.26 over the convex part,
at $t=0.21$. (b)~On the centre plane, the bands of $\eta_S$ vanish on $S$; its Hessian gives the normal lines (purple)
and its third derivative the shape operator $W$, drawn as osculating arcs at two points. Right: through $p$, $\eta_S$
is $\frac12t^2$ along the normal and flat along the surface.}
\label{fig:wallcoords}
\end{figure}

\paragraph{Computed example (\cref{fig:hero}).} The mesh extrudes a two-dimensional fastback profile with rounded edges
and a narrower cabin through an implicit function, marching cubes, Taubin smoothing and decimation, to 31.8k vertices;
the car is 4.5 long, and its centre-plane section stays within 0.016 of the profile. Non-lifting potential flow with
$U_\infty=(1,0,0)$ is solved by the method of fundamental solutions with 2,300 sources and 5,200 collocation points; on
a unit sphere its largest $C_p$ error is 0.037. Panel (b) and \cref{fig:wallcoords} use exact unsigned distances on the centre plane and
principal curvatures by quadric fitting. \Cref{fig:quadrature} takes 256 farthest-point anchors over 80,000 area-uniform samples,
with Voronoi area weights; the attention test reads $C_p$ at 4,096 samples, the front half up to nine times denser, with
a Gaussian similarity of length 0.6 at a query above the roof and 32 draws per density. Panel (c) moves the surface
along its normal by $h(x)=\sum\cos(k\cdot x+\phi)$, five shapes at each wavelength from 0.2 to 3 with amplitudes
0.003--0.04$R$, where $R=1.51$ is the area-weighted RMS radius; to first order the perturbed normals are $n-\nabla_Sh$.
The flow is re-solved on each shape and mapped back to $S$ by closest point. Field error is the area-weighted RMS of
the $C_p$ difference, and distances are Hausdorff distances between vertex sets in $x/R$ alone or in $(x/R,\nnT)$, the
$d$ of \cref{prop:sufficiency}. With normal lines the largest ratio of field error to distance is 0.567 over all 30
shapes; in position alone it is 1.76 at wavelength 3 and 24.0 at wavelength 0.2. This checks the Lipschitz case of
\cref{prop:sufficiency} empirically on one body; it is not a bound on $\omega$, a supremum over all admissible shapes.

\paragraph{Proof of \cref{lem:primitive}.} Since $d_S(x)^2=\inf_{y\in S}|x-y|^2=|x|^2-\sup_{y\in S}(2x\cdot y-|y|^2)$,
the function $\tfrac12|x|^2-\eta_S(x)=\sup_{y\in S}(x\cdot y-\tfrac12|y|^2)$ is a supremum of affine functions, hence
convex, and $\eta_S$ vanishes exactly on the closed set $S$. Where $S$ is $C^3$, take a local unit normal $n$ and the
signed distance $t$, so that $\eta_S=\tfrac12t^2$ and $\nabla t=n\circ\pi_S$ near $S$ \citep{federer1959curvature}. Then
$\nabla\eta_S=t\nabla t=x-\pi_S(x)$ and $\nabla^2\eta_S=\nabla t\,\nabla t^{\top}+t\,\nabla^2t$, which is $\nnT$ on $S$.
For $X$ tangent at a point of $S$, $\partial_Xt=0$, so there
$\partial_X\nabla^2\eta_S=(\partial_Xn)n^{\top}+n(\partial_Xn)^{\top}$, and since $n^{\top}\partial_Xn=0$, applying it
to $n$ gives $\partial_Xn=WX$, with $W$ the shape operator \citep{ambrosio1998curvature}. Replacing $n$ by $-n$ changes
neither $\eta_S$ nor its derivatives. In wall coordinates $x=p+t\,n(p)$ with $p\in S$, $dx=(I+tW)\,dp+n\,dt$, so the
Euclidean metric is $g_S\big((I+tW)\,\cdot,(I+tW)\,\cdot\big)+dt^2$, with $g_S$ the surface metric, and every
differential operator written in these coordinates has coefficients built from $g_S$ and $W$; surface operators
follow likewise from the closest-point map $x-\nabla\eta_S$ \citep{ruuth2008simple}. For the stability statements, if
$d_S(x)\le r-\varepsilon$ then $\hat d(x)\le d_S(x)+\varepsilon\le r$, and if $\hat d(x)\le r$ then
$d_S(x)\le r+\varepsilon$. Each inclusion uses one sign of the error: the first needs only $\hat d\le d_S+\varepsilon$
on $\{d_S\le r-\varepsilon\}$, the second only $\hat d\ge d_S-\varepsilon$ on $\{\hat d\le r\}$. The sublevel
filtrations are therefore $\varepsilon$-interleaved, their persistence diagrams
lie within bottleneck distance $\varepsilon$ \citep{cohensteiner2007stability}, and every feature of persistence above
$2\varepsilon$ is matched. For a smooth kernel $\phi$ supported in the unit ball and $\phi_r=r^{-3}\phi(\cdot/r)$,
$\nabla^k(\phi_r*e)=(\nabla^k\phi_r)*e$ and $\lVert\nabla^k\phi_r\rVert_{L^1}=r^{-k}\lVert\nabla^k\phi\rVert_{L^1}$, so
$|\nabla^k(\phi_r*e)|\le C_k\varepsilon/r^k$ with $C_k=\lVert\nabla^k\phi\rVert_{L^1}$ wherever $|e|\le\varepsilon$ on
the $r$-ball. Finally, a projector output $\hat P$ within $\delta<1$ of $\nnT$ in operator norm has a top eigenvector
$\hat n$ with $\sin\angle(\hat n,n)\le\delta/(1-\delta)$: writing $\hat n=n\cos\vartheta+u\sin\vartheta$ with $u\perp n$,
the eigen-equation gives $\hat\lambda\sin\vartheta=u^{\top}(\hat P-\nnT)\hat n$, with eigenvalue
$\hat\lambda\ge1-\delta$ \citep{davis1970rotation}. Normals therefore come at full resolution from the projector,
while curvature costs one more derivative. $\square$

\paragraph{Linearised flow.} For a thin body in a uniform stream of speed $V$ and Mach number $M$, the perturbation
potential obeys $(1-M^2)\varphi_{xx}+\varphi_{yy}=0$, with $\varphi_y=V\,dy_w/dx$ on the wall $y=y_w(x)$
\citep{liepmann1957gasdynamics}. For $M>1$, $p-p_\infty=\rho V^2\theta/\sqrt{M^2-1}$ with $\theta$ the local slope, and
along a convex wall $d\theta/ds=-\kappa$, so $dp/ds=-\rho V^2\kappa/\sqrt{M^2-1}$. For $M<1$, $p-p_\infty$ is
proportional, with factor $\rho V^2/\sqrt{1-M^2}$, to the Hilbert transform of the slope, and $dp/ds$ to the same
transform of the curvature, a convolution whose kernel decays only as the inverse distance. For $y_w=h\cos kx$ and
$M<1$, $\varphi=(Vh/\beta)\,e^{-\beta ky}\sin kx$ with $\beta=\sqrt{1-M^2}$, and the wall pressure coefficient is
$-(2hk/\beta)\cos kx$. As $h\to0$ with $hk$ fixed, the wall converges to a flat one in Hausdorff distance while the
pressure does not, so for a metric on position alone the modulus $\omega$ of \cref{prop:sufficiency} would not tend to
zero with $r$, and the bound could not shrink with reconstruction error; with normal lines included the distance is of
order $hk$, and $\omega(r)$ is of order $r/\beta$.

\paragraph{Proof of \cref{prop:sufficiency}.} Write $z=\mathrm{enc}(S)$ and $\hat S=\mathrm{dec}(z)$. For each token
value $z$, choose, measurably, an admissible shape $S_z$ with $\mathrm{enc}(S_z)=z$ whose distance to $\mathrm{dec}(z)$
is no larger than that of any admissible shape with the same tokens. $S$ itself is such a shape, so
$d(S_z,\hat S)\le d(S,\hat S)$. The predictor $g(z,c,x)=F(S_z,c)(\pi_{S_z}(x))$ reads only the tokens, the condition and
the query point, and since $\omega$ is nondecreasing, its definition and the triangle inequality give
\[
\lVert g(z,c,\cdot)-F(S,c)\rVert_S\;\le\;\omega\big(d(S,S_z)\big)\;\le\;\omega\big(d(S,\hat S)+d(\hat S,S_z)\big)
\;\le\;\omega\big(2d(S,\hat S)\big).
\]
Squaring and averaging over shapes and conditions bounds the mean squared error of $g$, with $x$ drawn from the area
measure of $S$, by $\E\,\omega(2d(S,\hat S))^2$, and the best predictor from the same inputs, $\E[f(x)\mid z,c,x]$, does
no worse. If no minimiser exists, a near-minimiser gives the same bound with $2d(S,\hat S)+\eta$ for any $\eta>0$, which
changes nothing when $\omega(r)\le Lr$. No regularity of $F$ in shape is used: $\omega$ is defined for every field, possibly
infinite, nondecreasing and zero at zero, since $d(S,S')=0$ means $S'=S$; when reconstruction is exact, $S_z=S$ and the
error is zero. The flow is solved only on admissible shapes, never on $\hat S$. $\omega$ measures how strongly the field
responds to a change of shape in the metric $d$, and through it the bound ties the accuracy attainable from the tokens
to reconstruction quality. For the RANS fields of \cref{sec:experiments} it is not known: it is finite wherever the field
depends continuously on shape, can be large near leading edges, need not tend to zero where separation switches with
shape, and can be estimated from pairs of simulated shapes at a shared condition. $\square$

\paragraph{The converse.} Fix a displacement $v$ and let $G_v(S)(x)=d_S(x+v)$ for $x\in S$. For any $S'$ and
$x'=\pi_{S'}(x)$, $|d_S(x+v)-d_{S'}(x'+v)|\le|d_S(x+v)-d_{S'}(x+v)|+|x-x'|\le2\,d(S,S')$, because distance functions
are 1-Lipschitz and differ everywhere by at most the Hausdorff distance, so the modulus of $G_v$ in
\cref{prop:sufficiency} is at most $\omega(r)=2r$. Where $S$ is smooth with curvature $\kappa$,
$G_v(S)(x)=(v^{\top}\nnT v)^{1/2}+O(\kappa|v|^2)$. The best prediction of $G_v$ from the tokens and $x$ is therefore,
for small $v$, a prediction of the normal line at $x$, and for larger $v$ one of the distance at $x+v$: an encoder whose
worst-case error over fields with $\omega(r)\le2r$ is small must reconstruct both pretraining targets.

\paragraph{The particle view.} A fluid particle moving along a wall streamline of the outer flow with speed $V$ and
unit direction $t$ has acceleration $V\,(dV/ds)\,t+V^2\,dt/ds$. On the surface,
$dt/ds=\kappa_g\,(n\times t)-(t^{\top}Wt)\,n$, with $\kappa_g$ the geodesic curvature of the streamline, because
$t\cdot n=0$ gives $n\cdot dt/ds=-t\cdot\partial_tn=-t^{\top}Wt$. Euler's equation $\rho\,a=-\nabla p$ then gives
$dp/ds=-\rho V\,dV/ds$ along the streamline, $\partial p/\partial(n\times t)=-\rho V^2\kappa_g$ across it within the
surface, and $\partial p/\partial n=\rho V^2\,t^{\top}Wt$ off the wall. The surface supplies $W$ and $n$; the flow
supplies $V$, $t$ and $\kappa_g$.

\paragraph{The field view.} The particle view applies Euler's equation in Lagrange's description of the flow,
following a particle; the field view applies Lagrangian mechanics in Euler's description, varying a potential at fixed
points, as electromagnetism varies the potentials $(\Phi,\mathbf A)$ in the action of
$\tfrac12(|\mathbf E|^2-|\mathbf B|^2)$, with $\mathbf E=-\nabla\Phi-\partial_\tau\mathbf A$ and
$\mathbf B=\nabla\times\mathbf A$ in natural units and $\tau$ for time \citep{landau1975classical}. At lowest order in
derivatives, locality and symmetry fix such an action up to one function. For irrotational barotropic flow with velocity
$\nabla\varphi$, let the Lagrangian density depend only on the first derivatives of $\varphi$ at the same point and be
invariant under shifts $\varphi\mapsto\varphi+\mathrm{const}$, rotations and Galilean boosts, where a boost of velocity
$u$ takes $\nabla\varphi$ to $\nabla\varphi-u$ and $\partial_\tau\varphi$ to
$\partial_\tau\varphi+u\cdot\nabla\varphi-\tfrac12|u|^2$. The only invariant of $(\partial_\tau\varphi,\nabla\varphi)$
is then $\partial_\tau\varphi+\tfrac12|\nabla\varphi|^2$, so the Lagrangian density is a function of it alone; since the
charge of the shift symmetry, the derivative of the Lagrangian density with respect to $-\partial_\tau\varphi$, is the
mass density $\rho$, the Lagrangian density is the pressure, tied to $\varphi$ by Bernoulli's law
\citep{bateman1929variational,seliger1968variational,greiter1989hydrodynamic,son2006general}:
\begin{equation}
\int\!\!\int_{\Omega}p\;d^3x\,d\tau,\qquad \frac{dp}{\rho}=-\,d\Big(\partial_\tau\varphi+\tfrac12|\nabla\varphi|^2\Big),
\label{eq:action}
\end{equation}
over the fluid $\Omega$. Within this class the equation of state, which relates $\rho$ to $p$, is the only free
function; the far stream and, for lifting bodies, a Kutta condition at sharp trailing edges must still be supplied.
Under variations of $\varphi$ that vanish far from the body and at the end times, but are free on $S$, the action
changes by
\[
\int\!\!\int_{\Omega}\big(\partial_\tau\rho+\nabla\cdot(\rho\nabla\varphi)\big)\,\delta\varphi\;d^3x\,d\tau
+\int\!\!\int_{S}\rho\,\partial_n\varphi\;\delta\varphi\;dA\,d\tau ,
\]
with $n$ pointing into the fluid. Stationarity gives mass conservation, which Noether's theorem
\citep{noether1918invariante} attaches to the shift symmetry, as it attaches charge conservation to the global phase
symmetry of a charged field; in a superfluid, $\varphi$ is $\hbar/m$ times that phase
\citep{greiter1989hydrodynamic,son2006general}. Stationarity also gives the wall condition $\partial_n\varphi=0$ as a
natural boundary condition, implied rather than imposed. In incompressible flow the Lagrangian density is
$-\tfrac12\rho|\nabla\varphi|^2$ up to a constant and a total time derivative, and its Euler--Lagrange equation is
Laplace's.

\paragraph{Locality, symmetry and the wall.} The Lagrangian density is local, and so is its dependence on the wall. In
steady flow, on the tube $0<t<\ell$ within the reach $\ell$ of $S$, with $t$ the signed distance as in the proof of
\cref{lem:primitive}, the wall coordinates turn it into the pressure times $\sqrt g$, with the wall fixed at $t=0$,
where the pressure now depends on $g^{ij}\partial_i\varphi\,\partial_j\varphi$,
$g=g_S\big((I+tW)\,\cdot,(I+tW)\,\cdot\big)+dt^2$ and $\sqrt g=\det(I+tW)\sqrt{g_S}$. There the shape enters only
through the surface metric $g_S$ and the shape operator $W$ at the foot point, which \cref{lem:primitive} reads from
$\eta_S$, and a change of shape confined to the tube, as in the proof of \cref{prop:outer}, becomes a change of metric
there; beyond the reach the coordinates fail, and the shape enters through the rest of the domain. The condition enters
only through the far field, $\varphi\to U_\infty\cdot x$, and the constants of the equation of state. Long-range
dependence is not in the action but in its solution: solving the local Euler--Lagrange equation spreads the influence
of each part of the wall over the whole flow in steady subsonic potential flow, through the Green's function of
\cref{sec:levels}, and over its downstream Mach cone in supersonic flow. This matches the division of work suggested
in \cref{sec:levels}, in which the action's local data are geometry and its long-range solution is what simulations
teach. Several symmetries of the action have counterparts in \areals{}, built in or taught by augmentation. The action
sees $S$ through its metric and area measure, which do not depend on how $S$ is parametrised or sampled, and
area-weighted attention makes \areals{} approximately invariant to sampling density, cutting the drift under a ninefold
change by \densityReduction{} (\cref{eq:attn}, \cref{app:stability}). Reversing the orientation,
$(n,t,W)\mapsto(-n,-t,-W)$ with $t$ the signed distance, changes neither $tW$ nor $dt^2$, hence neither $g$ nor the
action, so for a closed body, whose fluid side $S$ fixes, the field depends on $S$ only through quantities blind to
orientation, the derivatives of $\eta_S$ (\cref{lem:primitive}): the sign-free targets discard nothing the action uses,
and the normal sign flips of pretraining teach the network the same symmetry. Rotating body and stream together leaves
the action unchanged, so the condition enters only through the stream relative to the body, its speed and direction,
which is what the sensors must recover; rotation augmentation in pretraining teaches the geometric part of this
invariance, while field training is not rotation-augmented. Finally, if a rigid motion or reflection maps $S$ to itself
and also fixes the condition, the field inherits that symmetry whenever the stationary point is unique: a
mirror-symmetric wing has a mirror-symmetric pressure field at zero sideslip but in general not in sideslip, so the
symmetry maps of \cref{sec:priors} extend to fields under symmetric conditions, not in general.

\paragraph{Waves and Mach cones.} Write $\varphi$ again for the perturbation of a uniform stream $Vx$ of Mach number
$M$, as in the linearised flow above. Expanding the pressure to second order, with $d\rho/dp$ the inverse square of the
speed of sound, gives the Lagrangian density
\[
\tfrac12\rho_\infty\big[(M/V)^2(\partial_\tau\varphi+V\partial_x\varphi)^2-|\nabla\varphi|^2\big]
\]
up to a constant and a total derivative. The total derivative is idle in the bulk but not at the wall, where it supplies
the tangency condition $\partial_n\varphi=-Vn_x$, linearised $\varphi_y=V\,dy_w/dx$: the body enters through its slope.
The density is a squared time derivative minus a squared gradient, the action of a massless scalar field, whose wave
equation each component of $\mathbf A$ obeys in the Lorenz gauge; here its Euler--Lagrange equation is the convected
wave equation. In steady flow it is $\tfrac12\rho_\infty[(M^2-1)\varphi_x^2-|\nabla_{\!\perp}\varphi|^2]$, with
$\nabla_{\!\perp}$ the gradient across the stream, and its Euler--Lagrange equation is the linearised equation above.
The signature of this form decides the domain of dependence. For $M>1$ it is Lorentzian, with $x$ as time and the Mach
lines $dy/dx=\pm1/\sqrt{M^2-1}$ as light cones: in two dimensions the solution above the wall is a function of
$x-\sqrt{M^2-1}\,y$ alone, a wave carried along Mach lines, which gives the local law
$p-p_\infty=\rho V^2\theta/\sqrt{M^2-1}$, and in three the wall pressure at a point depends only on the wall within its
upstream Mach cone, all of which contributes. For $M<1$ the form is definite and the equation elliptic, and every part
of the wall contributes, through the Hilbert transform in two dimensions.

\paragraph{Vorticity.} Both views extend to the vortices that \cref{prop:outer} excludes. In Lagrange's description,
Hamilton's principle for the fluid particles is invariant under relabelling particles, and this symmetry yields
Kelvin's circulation theorem \citep{bretherton1970note,salmon1988hamiltonian}: in barotropic flow, fluid that arrives
irrotational stays so outside the boundary layer, the wake and the region behind curved shocks, which is why the outer
flow has a potential there; the same principle moves the vorticity that is shed. In Euler's description, a variational
principle still gives the velocity $\nabla\times\Psi$ that a given vorticity $\omega$ induces, with $\Psi$ stationary
for the density $\tfrac12|\nabla\times\Psi|^2-\omega\cdot\Psi$, the magnetostatic density up to sign with $\omega$ in
place of the current; the velocity kernel it yields is the Biot--Savart kernel, which decays only as the inverse square
of distance, so a vortex acts on the wall at long range, as a current acts through its magnetic field. Evolving
$\omega$ in this description needs Clebsch potentials \citep{seliger1968variational}. Where the vortex sits depends on
where the flow separates, which neither principle determines; at sharp edges, such as a delta wing's leading edges, a
Kutta condition supplies it.

\begin{proposition}[Stability of the outer flow]
\label{prop:outer}
Let $S$ bound a smooth body of reach $\ell$ in potential flow without circulation, and let $S'=\{p+h(p)\,n(p)\}$ with
$|h|\le\varepsilon$, normals within $\delta$ and shape operators within $\varepsilon_W$. Mapped back along the wall coordinates
of \cref{lem:primitive}, the velocity changes in energy by at most $C|U_\infty|(\varepsilon/\ell+\delta)$ and the wall
pressure, in the norm of \cref{prop:sufficiency}, by at most $C\rho|U_\infty|^2(\varepsilon/\ell+\delta+\ell\varepsilon_W)$;
in linearised flow the curvature term drops out.
\end{proposition}

\paragraph{Proof of \cref{prop:outer}.} Write $\varphi=U_\infty\cdot x+\psi$ with $\Delta\psi=0$ in the fluid,
$\partial_n\varphi=0$ on $S$ and $\nabla\psi\to0$ at infinity, and $\varphi'$ likewise for $S'$. Let $t$ be the signed
distance to $S$, positive in the fluid, and $\chi$ a smooth cut-off with $\chi(0)=1$ and $\chi(s)=0$ for
$|s|\ge\frac12$. The map $T(x)=x+\chi(t/\ell)\,h(\pi_S(x))\,n(\pi_S(x))$ is the identity outside the tube
$\{|t|<\ell/2\}$, and for small $\varepsilon/\ell$ and $\delta$ it is a diffeomorphism taking $S$ to $S'$ and the
fluid around $S$ to the fluid around $S'$. Since $\nabla t=n\circ\pi_S$ and $\lVert\partial(n\circ\pi_S)\rVert\le2/\ell$
in the tube, $\lVert DT-I\rVert_\infty\le C(\varepsilon/\ell+\delta)$, and the second derivatives of $T$ are at most
$C(\varepsilon_W+\delta/\ell+\varepsilon/\ell^2)$. The pulled-back potential $\tilde\varphi=\varphi'\circ T$ solves
$\nabla\cdot(A\nabla\tilde\varphi)=0$ with $A\nabla\tilde\varphi\cdot n=0$ on $S$, where
$A=\det(DT)\,DT^{-1}DT^{-\top}$ differs from $I$ only in the tube, by at most $C(\varepsilon/\ell+\delta)$. In the
field view, $A=\sqrt{\det g_T}\,g_T^{-1}$ for the metric $g_T=DT^{\top}DT$ that $T$ pulls back, so this is the
Euler--Lagrange equation, with its natural boundary condition, of the incompressible action of the flow around $S'$
carried onto the fluid around $S$: the change of shape has become a change of metric, set to first order by the errors
in position and normal line, the two pretraining targets, while the error in curvature enters only through derivatives
of the metric. The
difference $w=\tilde\varphi-\varphi$ decays at infinity and satisfies
$\int\nabla w\cdot\nabla v=-\int(A-I)\nabla\tilde\varphi\cdot\nabla v$ for every test function $v$, so
$\lVert\nabla w\rVert_{L^2}\le\lVert A-I\rVert_\infty\lVert\nabla\tilde\varphi\rVert_{L^2(\text{tube})}\le
C|U_\infty|(\varepsilon/\ell+\delta)$, the energy bound \citep{sokolowski1992shape}. For wall values, $w$ solves a
Neumann problem whose source $-\nabla\cdot\big((A-I)\nabla\tilde\varphi\big)$ and boundary data
$-(A-I)\nabla\tilde\varphi\cdot n$ involve the first derivatives of $A$, hence the curvature term; the $H^2$ estimate
for $C^{1,1}$ boundaries \citep{grisvard1985elliptic} and the trace theorem bound $\nabla w$ on $S$ in $L^2$ by
$C|U_\infty|(\varepsilon/\ell+\delta+\ell\varepsilon_W)$. The wall velocity $(\nabla\varphi')\circ T=DT^{-\top}\nabla\tilde\varphi$
then differs from $\nabla\varphi$ by at most $|DT^{-\top}-I|\,|\nabla\tilde\varphi|+|\nabla w|$, Bernoulli's law
$p=p_\infty+\frac12\rho(|U_\infty|^2-|\nabla\varphi|^2)$ turns this into the pressure bound, and replacing $T(p)$ by the
closest point $\pi_{S'}(p)$ changes the comparison by $O(\varepsilon\delta)$. The constants depend on $S$ and on the
curvature bound for $S'$ through the regularity of the two solutions. In linearised two-dimensional flow the wall
pressure is $\rho V^2/\sqrt{1-M^2}$ times the Hilbert transform of the slope, an isometry of $L^2$, so the bound holds
with the slope error alone. With circulation, the Kutta condition fixes it at the trailing edge; in thin-airfoil
theory the circulation depends on the camber slope through a bounded weight, but sharp edges fall outside this proof.
$\square$

\paragraph{Fine-tuning and the prior.} Write a field model as $p_\theta(f\mid S,\xi)$, where $\xi$ is what the model
is told: nothing, the condition, or sensor values. Let $\theta_{\mathrm{geo}}$ be the weights of the network pretrained to complete
raw surfaces, with new field parameters at their initial values. Pretraining fixes the centre of the prior
$p_{\mathrm{geo}}(\theta)=\mathcal{N}(\theta_{\mathrm{geo}},\mathrm{diag}(\alpha^2))$ before any simulation is seen,
and simulations $\mathcal{D}=\{(S_i,\xi_i,f_i)\}_{i=1}^{N}$ update it:
\begin{equation}
p(\theta\mid\mathcal{D})\;\propto\;p_{\mathrm{geo}}(\theta)\prod_{i=1}^{N}p_\theta(f_i\mid S_i,\xi_i).
\label{eq:prior}
\end{equation}
With $p_\theta(f\mid S,\xi)=\mathcal{N}(f_\theta(S,\xi),\sigma^2I)$, the negative logarithm of \cref{eq:prior} is, up
to a constant,
\[
\frac{1}{2\sigma^2}\sum_{i=1}^{N}\big\lVert f_i-f_\theta(S_i,\xi_i)\big\rVert^2
+\sum_{k}\frac{\lVert\theta_k-\theta_{\mathrm{geo},k}\rVert^2}{2\alpha_k^2},
\]
with $k$ running over parameter groups. Adding the second term to the fine-tuning loss gives L2-SP
\citep{li2018l2sp}, whose minimiser is the mode of \cref{eq:prior}. Plain fine-tuning regularises implicitly: for least
squares in a model linear in $\theta$, gradient flow from $\theta_{\mathrm{geo}}$ stopped at time $t$ has risk within a
factor of about 1.69 of ridge regression towards $\theta_{\mathrm{geo}}$ with penalty $1/t$ \citep{ali2019continuous}.
Early stopping therefore acts like a Gaussian prior around $\theta_{\mathrm{geo}}$ whose width grows with training time,
and a smaller learning rate for a parameter group acts like a tighter prior on it. For our AdamW-trained network the correspondence is a heuristic, and AdamW's weight decay adds a pull
towards zero.

\paragraph{A conjugate last layer.} Fix the trunk at $\theta_{\mathrm{geo}}$, let $\phi(x)\in\mathbb{R}^m$ be the
decoder's features at query $x$, which at $\theta_{\mathrm{geo}}$ depend only on the shape, since the new input
columns start at zero, and put a Gaussian prior
$w\sim\mathcal{N}(w_0,\Sigma)$ on the linear field head, centred on its initial value, so that $f(x)=\phi(x)^{\top}w$
plus Gaussian noise of variance $\sigma^2$. Stacking the features of the simulated points into $\Phi$ and their values
into $\mathbf{f}$, the posterior over $w$ is Gaussian with precision $P=\Sigma^{-1}+\Phi^{\top}\Phi/\sigma^2$ and mean
$P^{-1}(\Sigma^{-1}w_0+\Phi^{\top}\mathbf{f}/\sigma^2)$, and the evidence is
$\mathcal{N}(\mathbf{f};\Phi w_0,\Phi\Sigma\Phi^{\top}+\sigma^2I)$ \citep{mackay1992bayesian,snoek2015scalable}. The
prior over field models is then conjugate to the simulations, and the evidence under pretrained and under random
features can be compared directly.

\paragraph{A prior fixed before the simulations.} Geometry pretraining never sees a field, and its meshes are
drawn independently of the simulated geometries, so the centre of $p_{\mathrm{geo}}$ is chosen independently of
$\mathcal{D}$, the condition PAC-Bayes bounds place on a prior
\citep{mcallester1999pacbayes,dziugaite2017nonvacuous}. For independently drawn geometries, a bounded loss and any
distribution $Q$ over $\theta$, they bound the expected loss of the stochastic predictor $Q$ on new geometries by its
training value plus a term that grows with $\mathrm{KL}(Q\,\Vert\,p_{\mathrm{geo}})$, provided the widths $\alpha$ are
also fixed in advance.

\section{Expanded Benchmarks}
\label{app:expanded}

\begin{table}[H]
\caption{Expanded benchmarks, three seeds: hidden-region nRMSE ($\downarrow$) at each sensor count. Shape only and the
true-condition reference see no sensors; scratch is the sensor model from random initialisation. The last column is the
95\% geometry-grouped bootstrap interval of the paired difference between \areals{} and scratch.}
\label{tab:expanded}
\centering
\small
\setlength{\tabcolsep}{4pt}
\begin{tabular}{@{}llrccccc@{}}
\toprule
Benchmark & Test set & $K$ & Shape only & Scratch & \areals{} & True condition & \areals{} $-$ scratch \\
\midrule
DoubleDelta & 16 wings, $18$--$19^\circ$ & 5 & 0.373 & 0.409 & 0.310 & 0.193 & $[-0.157,-0.046]$ \\
 &  & 20 & 0.373 & 0.409 & 0.206 & 0.193 & $[-0.235,-0.170]$ \\
 &  & 100 & 0.375 & 0.411 & 0.178 & 0.194 & $[-0.258,-0.210]$ \\
 &  & 500 & 0.383 & 0.420 & 0.187 & 0.200 & $[-0.258,-0.208]$ \\
\midrule
BlendedNet++ & 1,872 aircraft, high Mach & 5 & 0.742 & 0.281 & 0.240 & 0.066 & $[-0.044,-0.037]$ \\
 &  & 20 & 0.743 & 0.212 & 0.176 & 0.066 & $[-0.037,-0.034]$ \\
 &  & 100 & 0.745 & 0.191 & 0.145 & 0.067 & $[-0.046,-0.044]$ \\
 &  & 500 & 0.755 & 0.189 & 0.138 & 0.068 & $[-0.052,-0.050]$ \\
\midrule
DrivAerML & 50 official test cars & 5 & 0.240 & 0.294 & 0.241 & 0.240 & $[-0.056,-0.050]$ \\
 &  & 20 & 0.240 & 0.295 & 0.242 & 0.241 & $[-0.056,-0.050]$ \\
 &  & 100 & 0.242 & 0.296 & 0.243 & 0.242 & $[-0.056,-0.050]$ \\
 &  & 500 & 0.248 & 0.303 & 0.249 & 0.248 & $[-0.057,-0.051]$ \\
\midrule
DrivAerNet++ & 848 test cars & 5 & 0.204 & 0.208 & 0.204 & 0.204 & $[-0.004,-0.003]$ \\
 &  & 20 & 0.204 & 0.208 & 0.204 & 0.204 & $[-0.004,-0.003]$ \\
 &  & 100 & 0.204 & 0.208 & 0.204 & 0.204 & $[-0.004,-0.003]$ \\
 &  & 500 & 0.204 & 0.207 & 0.204 & 0.204 & $[-0.004,-0.003]$ \\
\bottomrule
\end{tabular}
\end{table}

\begin{table}[H]
\caption{Gaussian fields: share of the exact posterior mean's gain over the prior recovered (\%, $\uparrow$),
averaged over 5, 20, 100 and 500 sensors and 60\% hidden patches; 16 held-out families, eight fields each, three seeds.
With the trunk frozen, pretraining adds 25.7 points (95\% interval 21.6--29.5).}
\label{tab:posterior}
\centering
\small
\begin{tabular}{@{}lcc@{}}
\toprule
Geometry trunk & Frozen & Fine-tuned \\
\midrule
Random initialisation & 4.7 & 25.4 \\
\areals{} pretraining & \textbf{30.4} & \textbf{82.6} \\
\bottomrule
\end{tabular}
\end{table}

\begin{table}[H]
\caption{Gaussian-field efficiency by observation setting with a frozen geometry trunk (\%, $\uparrow$): the share of the
exact posterior mean's improvement over the prior that each network recovers. Fine-tuned means are in
\cref{tab:posterior}.}
\label{tab:gaussian-full}
\centering
\small
\begin{tabular}{@{}lcc@{}}
\toprule
Observations & Random trunk & \areals{} pretraining \\
\midrule
5 sensors & $-0.2$ & 6.9 \\
20 sensors & 0.8 & 18.5 \\
100 sensors & 2.8 & 35.2 \\
500 sensors & 9.3 & 54.8 \\
60\% hidden patches & 10.7 & 36.3 \\
\midrule
Mean & 4.7 & 30.4 \\
\bottomrule
\end{tabular}
\end{table}

\begin{table}[H]
\caption{Masked against unmasked geometry pretraining, transferred to pressure. Hidden-region nRMSE ($\downarrow$) for
three geometry seeds (17, 18, 19), each at epoch 11 (10,626 updates) and fine-tuned once with the field recipe below;
$K=0$ observes nothing. Intervals jointly resample seeds and geometry families (5,000 draws). The 20-sensor endpoint
was specified before results and uses a 97.5\% interval; the others use 95\%.}
\label{tab:mask}
\centering
\small
\setlength{\tabcolsep}{4pt}
\begin{tabular}{@{}lrccc@{}}
\toprule
Benchmark & $K$ & Masked & Unmasked & Masked $-$ unmasked \\
\midrule
DoubleDelta, $18$--$19^\circ$ & 0 & 0.4306 & 0.4211 & $+0.0095$ $[-0.0043, +0.0277]$ \\
 & 5 & 0.3119 & 0.3036 & $+0.0083$ $[-0.0059, +0.0223]$ \\
 & 20 & 0.2081 & 0.2057 & $+0.0025$ $[-0.0086, +0.0139]$ \\
 & 100 & 0.1782 & 0.1801 & $-0.0018$ $[-0.0117, +0.0072]$ \\
 & 500 & 0.1867 & 0.1847 & $+0.0020$ $[-0.0130, +0.0156]$ \\
\midrule
DrivAerML, official & 0 & 0.2403 & 0.2412 & $-0.0009$ $[-0.0041, +0.0028]$ \\
 & 5 & 0.2403 & 0.2413 & $-0.0010$ $[-0.0042, +0.0028]$ \\
 & 20 & 0.2406 & 0.2415 & $-0.0009$ $[-0.0044, +0.0034]$ \\
 & 100 & 0.2418 & 0.2428 & $-0.0010$ $[-0.0042, +0.0027]$ \\
 & 500 & 0.2477 & 0.2487 & $-0.0010$ $[-0.0042, +0.0027]$ \\
\bottomrule
\end{tabular}
\end{table}

\paragraph{Protocol.} Fields are the datasets' native nondimensional pressure coefficients on 32,000 surface points,
with 16,384 geometry samples per case. Queries are a fixed uniform subset of native surface vertices with their lumped
area weights, 2,048 per example. Field training uses batch 8, 40 epochs, learning rate $10^{-3}$ for all parameters,
weight decay 0.05 and bfloat16, with 256 tokens and 4,096 encoder points; 10\% of examples observe nothing and 90\%
observe scattered sensors. Normalisation uses training records only. Checkpoints are selected by mean validation nRMSE
over the four sensor counts on at most 64 development cases, and each test set is scored once with the selected
checkpoint. Sensor locations are fixed independently of the pressure labels and shared by all methods. BlendedNet++
splits by geometry hash and Mach number: training uses Mach up to the 70th percentile, validation the band to the 85th
and testing above it, with disjoint geometries. DrivAerNet++ is deduplicated before an 80/10/10 geometry-hash split.

\paragraph{Gaussian fields.} Fresh Gaussian random fields are drawn on 16 held-out geometry families, eight per
geometry. Field adaptation runs 9,120 updates with the geometry trunk frozen or trainable. Efficiency is the share of
the exact posterior mean's improvement over the prior recovered by the network, averaged with equal weight over five
observation settings; intervals jointly resample seeds and geometries.

\begin{table}[H]
\caption{Geometry pretraining against the same architecture from scratch: hidden-region nRMSE ($\downarrow$) at 20
sensors unless marked; three seeds, matched examples, updates and learning rate; every interval excludes zero.}
\label{tab:pretrain}
\centering
\small
\setlength{\tabcolsep}{4pt}
\begin{tabular}{@{}lrccc@{}}
\toprule
Benchmark & Train & Scratch & \areals{} & Reduction \\
\midrule
DoubleDelta, $18$--$19^\circ$ & 228 & 0.409 & \textbf{0.206} & 49.6\% \\
BlendedNet++, high Mach & 8,736 & 0.212 & \textbf{0.176} & 16.8\% \\
DrivAerML, official & 400 & 0.295 & \textbf{0.242} & 18.0\% \\
DrivAerML, reserved$^\ast$ & 420 & 0.284 & \textbf{0.226} & 20.2\% \\
\bottomrule
\end{tabular}
\par\smallskip{\footnotesize $^\ast$48 reserved cars, averaged over all 25 observation settings.}
\end{table}

\paragraph{Simulation efficiency.} Each fit trains the pretrained model, or the same architecture from scratch, on a
sixteenth, an eighth, a quarter, a half or all of a dataset's training simulations for 8,000 updates, keeps the
checkpoint with the lowest validation error for each sensor setting, and is scored once on the dataset's test
simulations, which are common to every budget and model. WindsorML is split into 274 training, 35 validation and 35
test geometries, DrivAerML uses its official 400/34/50 split and DrivAerNet++ 6,492, 778 and 848. A budget meets the
target when the mean test nRMSE of its three seeds is at most 0.25, and the arrows of \cref{fig:efficiency} join the
smallest tested budgets that do; counts are not interpolated. Intervals come from a paired bootstrap over seeds and
test geometries with 5,000 draws, and every paired interval for the difference between \areals{} and scratch excludes
zero. Savings with validation add the validation simulations each model uses, and exclude the test simulations and
the compute of geometry pretraining; they are 59.3\%, 46.1\% and 40.3\% on WindsorML, DrivAerML and DrivAerNet++. With 20 sensors the smallest budgets are the same. At a target of 0.3 they are 18
against 69, 100 against 200 and 406 against 812, \areals{}'s being the smallest tested on WindsorML and DrivAerNet++;
at 0.2, \areals{} needs 69 WindsorML simulations against 274, on DrivAerML only \areals{} reaches it, and on
DrivAerNet++ neither does. Fits still improving late in training were flagged: none of 30 on WindsorML, 6 of 30 on
DrivAerML and 17 of 30 on DrivAerNet++. The gain is largest where simulations are scarcest: at 8,000 updates per fit it is 36--39\% with 18--35
WindsorML simulations, 25--27\% with up to 100 DrivAerML simulations and 8.7\% with 406 DrivAerNet++ simulations,
against 22\%, 17.5\% and 6.4\% with the full training sets. An independent replay verified all three comparisons.

\section{DoubleDelta Results and Training Budgets}
\label{app:dd}

\paragraph{From-scratch baseline.} The same sensor transformer trained from random initialisation on the nine-angle
split, with identical observation sets, learning rates and updates, scores 0.2219 at 20 sensors and 0.2177 at 500,
against 0.1637 and 0.1410 for \areals{}, means of three fine-tuning seeds (\cref{tab:dd-full}): geometry pretraining
lowers the error by 26.2\% and 35.2\%.

\begin{figure}[H]
\centering
\begin{minipage}{0.5\textwidth}
\centering
\begin{tikzpicture}
\begin{axis}[
    width=\columnwidth, height=5.3cm,
    xmode=log, ymode=log,
    xmin=4, xmax=640, ymin=0.12, ymax=1.05,
    xtick={5,20,100,500}, xticklabels={5,20,100,500},
    ytick={0.15,0.3,0.6,0.9}, yticklabels={0.15,0.3,0.6,0.9},
    minor tick num=0,
    xlabel={Pressure sensors $K$},
    ylabel={Hidden-region nRMSE},
    tick label style={font=\scriptsize},
    label style={font=\small},
    axis line style={arealmuted},
    tick style={arealmuted},
    ymajorgrids, grid style={arealgrid},
    legend cell align=left,
    legend columns=2,
    legend style={font=\scriptsize, draw=none, fill=none, at={(0.5,1.02)}, anchor=south,
                  column sep=6pt},
    clip=false,
  ]
  \addplot[color=arealmuted, line width=1.1pt, mark=square*, mark size=1.8pt]
    coordinates {(5,0.302) (20,0.302) (100,0.304) (500,0.310)};
  \addlegendentry{geometry only}
  \addplot[color=arealteal, line width=1.6pt, mark=*, mark size=2.1pt]
    coordinates {(5,0.2111) (20,0.1637) (100,0.1434) (500,0.1410)};
  \addlegendentry{sensors}
  \addplot[color=arealfg, line width=1.0pt, dashed, mark=none]
    coordinates {(5,0.137) (20,0.137) (100,0.138) (500,0.143)};
  \addlegendentry{true angle}
  \addplot[color=arealwarn, line width=1.2pt, mark=triangle*, mark size=2pt]
    coordinates {(5,0.7451882254) (20,0.6422355529) (100,0.5366860914) (500,0.5146918886)};
  \addlegendentry{GABI}
  \draw[arealteal, line width=0.7pt, |-|] (axis cs:24,0.296) -- (axis cs:24,0.140);
  \node[font=\scriptsize, text=arealteal, anchor=west, align=left] at (axis cs:26,0.205)
    {$G=\gapClosed{}$ of the MSE gap\\closed at $K=20$};
\end{axis}
\end{tikzpicture}
\end{minipage}
\caption{\textbf{Sparse measurements recover the unreported condition.} On 16 test wings at nine angles, the
\areals{} measurement model approaches the true-angle reference in one pass. GABI uses its published schedule,
50,000 prior samples per geometry and graphs built along the mesh. \areals{} and GABI are means of three seeds; $G$ is
for the first \areals{} seed.}
\label{fig:doubledelta}
\end{figure}
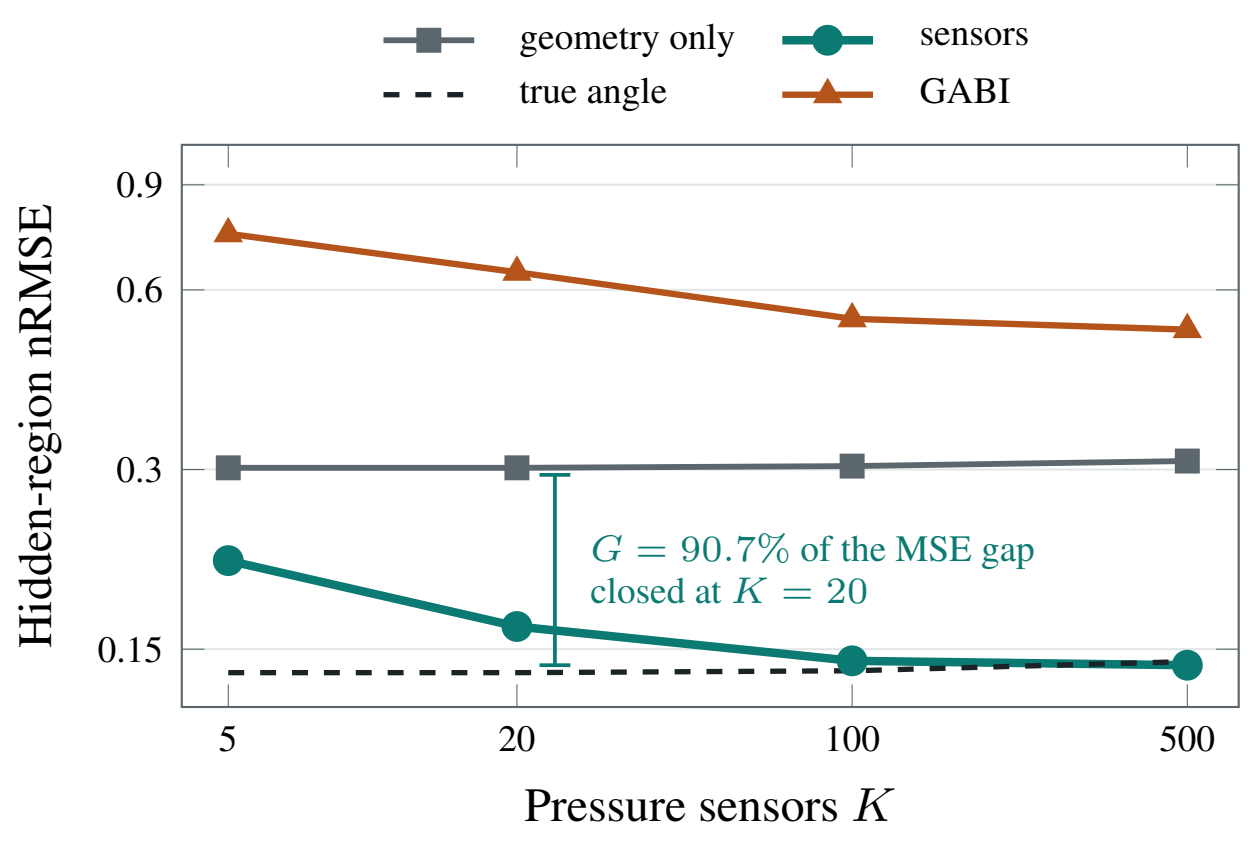

\paragraph{Fine-tuning seeds.} The nine-angle \areals{} model is fine-tuned from the one production checkpoint with
three seeds, each for 40 epochs (10,800 updates at batch 8) and selected by the same rule on 36 development cases. Seeds
1 and 2 were fixed before they ran but after the first seed's test results were known, and nothing was tuned
afterwards. The seeds share one geometry checkpoint, so their spread, 0.003 nRMSE at 20 sensors, measures fine-tuning
variation, not pretraining variation.

\paragraph{Explicit condition inference.} This baseline evaluates \cref{eq:posterior} with the true-condition model of
the extrapolation split, which is trained with condition labels on $11$--$16^\circ$. Over the conditions of the training
range, the support of the prior $p(c)$, it takes the posterior mode (plug-in) or averages the model's predictions under
the posterior (mixture, with its settings selected on the validation wings); the two agree to five decimals. At
$18$--$19^\circ$ the posterior cannot reach the true angle, and the error stays at 0.300--0.307 from 20 to 500 sensors,
against 0.206, 0.178 and 0.187 for \areals{}; at each of these sensor counts the two methods' 95\% bootstrap intervals
do not overlap (\cref{tab:explicit}). Told the true angle, the same network reaches 0.193--0.200 (\cref{tab:expanded}),
so a search widened beyond the training range could at best approach that, and \areals{} is below it from 100 sensors.
Within the training range explicit inference can approach the true-condition reference, but it needs condition labels
to train and a search at test time; \areals{} needs neither. Timing is the median over three seeds, four validation
wings and three repeats, with one warm-up excluded, on one B200 in FP32, and covers encoding, search and full-field
prediction but not geometry preprocessing or checkpoint loading.

\begin{table}[h]
\caption{Explicit condition inference against \areals{} with the condition unreported: hidden-region nRMSE
($\downarrow$), mean $\pm$ s.d.\ over three seeds, with 95\% bootstrap intervals and the median time per case.
Explicit inference evaluates \cref{eq:posterior} with the true-condition model over the training conditions; the
mixture is shown, and the plug-in estimate agrees to five decimals. --- marks entries not evaluated.}
\label{tab:explicit}
\centering
\small
\setlength{\tabcolsep}{4pt}
\begin{tabular}{@{}lrcccc c@{}}
\toprule
 & & \multicolumn{2}{c}{\areals{}} & \multicolumn{2}{c}{Explicit inference} & Time per case (s) \\
\cmidrule(lr){3-4}\cmidrule(lr){5-6}
Benchmark & $K$ & nRMSE & 95\% interval & nRMSE & 95\% interval & \areals{} / explicit \\
\midrule
DoubleDelta, $18$--$19^\circ$ & 5 & 0.310 & --- & --- & --- & --- \\
 & 20 & \textbf{0.2060}\,{\scriptsize$\pm$0.0035} & [0.181, 0.230] & 0.2999\,{\scriptsize$\pm$0.0007} & [0.264, 0.336] & 0.122 / 0.356 \\
 & 100 & \textbf{0.1775}\,{\scriptsize$\pm$0.0061} & [0.154, 0.202] & 0.2999\,{\scriptsize$\pm$0.0023} & [0.263, 0.337] & 0.123 / 0.355 \\
 & 500 & \textbf{0.1872}\,{\scriptsize$\pm$0.0086} & [0.163, 0.211] & 0.3066\,{\scriptsize$\pm$0.0002} & [0.268, 0.345] & 0.123 / 0.354 \\
\midrule
DoubleDelta, nine angles & 20 & 0.1637\,{\scriptsize$\pm$0.0030} & --- & --- & --- & --- \\
BlendedNet++, high Mach & 20 & 0.176 & --- & --- & --- & --- \\
\bottomrule
\end{tabular}
\end{table}

\begin{table}[h]
\caption{DoubleDelta on the 144 test cases (16 physical wings, nine angles). Hidden-region nRMSE ($\downarrow$).
Geometry-only and true-angle predictions are fixed for a case; their scores change with $K$ because the hidden scoring
region excludes each setting's observed points. Sensors are sampled by area, so 500 sensors remove about 10\% of
the area from the scored region. Measurements: mean of three fine-tuning seeds, with standard deviations of at most
0.0043; the two references are single models.}
\label{tab:dd-full}
\centering
\small
\begin{tabular}{@{}rccc@{}}
\toprule
Sensors $K$ & Geometry only & Measurements & True angle \\
\midrule
5 & 0.3021 & 0.2111 & 0.1373 \\
20 & 0.3023 & 0.1637 & 0.1374 \\
100 & 0.3036 & 0.1434 & 0.1383 \\
500 & 0.3101 & 0.1410 & 0.1426 \\
\bottomrule
\end{tabular}
\end{table}

\begin{table}[h]
\caption{Training budgets on the nine-angle DoubleDelta split. GABI's published schedule (10,000 updates at batch
100, here 455 epochs) gives about as many optimiser updates as each \areals{} model and 11.375 times as many
field-example exposures. Geometry-only pretraining is additional to the field-training counts. GABI reconstructs the
full 32,000-point field; \areals{} samples 2,048 queries per masked example. GABI's checkpoint minimises development
reconstruction plus MMD loss.}
\label{tab:budgets}
\centering
\small
\begin{tabular}{@{}lrr@{}}
\toprule
Budget & \areals{}, per model & GABI, published schedule \\
\midrule
Epochs & 40 & 455 \\
Batch size & 8 & 100 \\
Optimiser updates & 10,800 & 10,010 \\
Field-example exposures & 86,400 & 982,800 \\
Training seeds & 3 & 3 per graph \\
\bottomrule
\end{tabular}
\end{table}

\begin{table}[h]
\caption{GABI and \areals{} on the same 144 DoubleDelta \emph{test} cases with the angle unreported: hidden-region
nRMSE and integrated-force error ($\downarrow$), means of three seeds; GABI uses graphs built along the mesh. The
true-angle model's force error is 1.9\% at 20 sensors and 1.8\% at 500. At 20 sensors, mean~$\pm$~s.d.\ over seeds,
\areals{} has 0.164~$\pm$~0.003 nRMSE and 3.7~$\pm$~0.4\% force error, against 0.642~$\pm$~0.155 and 14.7~$\pm$~1.7\% for
GABI. With nothing observed, GABI's error is already 2.9 times that of \areals{}; 20 sensors then remove 25\% of
GABI's error and 45\% of that of \areals{}.}
\label{tab:gabi}
\centering
\small
\setlength{\tabcolsep}{5pt}
\begin{tabular}{@{}rcccc@{}}
\toprule
 & \multicolumn{2}{c}{nRMSE} & \multicolumn{2}{c}{Force error (\%)} \\
\cmidrule(lr){2-3}\cmidrule(l){4-5}
Sensors $K$ & \areals{} & GABI & \areals{} & GABI \\
\midrule
0 & \textbf{0.300} & 0.858 & \textbf{15.5} & 25.3 \\
5 & \textbf{0.211} & 0.745 & \textbf{8.6} & 21.4 \\
20 & \textbf{0.164} & 0.642 & \textbf{3.7} & 14.7 \\
100 & \textbf{0.143} & 0.537 & \textbf{2.2} & 12.2 \\
500 & \textbf{0.141} & 0.515 & \textbf{1.5} & 10.6 \\
\bottomrule
\end{tabular}
\end{table}

\begin{table}[h]
\caption{GABI and \areals{} on identical test observation sets: hidden-region nRMSE, means of three seeds with their
standard deviations. GABI runs its published schedule, with each point joined to its eight nearest vertices by
straight-line distance or by shortest distance along mesh edges. The last two columns give GABI's along-mesh minus
straight-line difference and the along-mesh GABI minus \areals{} difference, each with a 95\% interval that resamples
the 16 physical wings, keeping all angles and seeds, conditional on the trained models. The geometry-only and
true-angle references are single models. Patch settings use 64 cells. At no observations, \areals{} uses the
measurement model with zero presence flags; GABI averages decoded draws from its unconditional latent prior. This is
the error of the \emph{prior mean}, not the mean error of individual prior draws.}
\label{tab:gabi-full}
\centering
\footnotesize
\setlength{\tabcolsep}{3pt}
\newcommand{\gabiHd}[1]{\begin{tabular}{@{}c@{}}#1\end{tabular}}
\begin{tabular}{@{}lccccccc@{}}
\toprule
 & \multicolumn{3}{c}{\areals{}} & \multicolumn{3}{c}{GABI} & \\
\cmidrule(lr){2-4}\cmidrule(lr){5-7}
Observed & \gabiHd{Geometry\\only} & \gabiHd{Measurements} & \gabiHd{True\\angle} & \gabiHd{Straight\\line} & \gabiHd{Along\\mesh} &
\gabiHd{Along mesh $-$\\straight line} & \gabiHd{GABI $-$\\\areals{}} \\
\midrule
Nothing & 0.302 & 0.300\,{\scriptsize$\pm$0.001} & 0.137 & 0.914 & 0.858\,{\scriptsize$\pm$0.037} & $-0.056$ {\scriptsize[$-0.088$, $-0.027$]} & 0.558 {\scriptsize[0.422, 0.696]} \\
5 sensors & 0.302 & 0.211\,{\scriptsize$\pm$0.004} & 0.137 & 0.834 & 0.745\,{\scriptsize$\pm$0.104} & $-0.089$ {\scriptsize[$-0.160$, $-0.015$]} & 0.534 {\scriptsize[0.447, 0.617]} \\
20 sensors & 0.302 & 0.164\,{\scriptsize$\pm$0.003} & 0.137 & 0.736 & 0.642\,{\scriptsize$\pm$0.155} & $-0.094$ {\scriptsize[$-0.156$, $-0.027$]} & 0.479 {\scriptsize[0.402, 0.556]} \\
100 sensors & 0.304 & 0.143\,{\scriptsize$\pm$0.001} & 0.138 & 0.648 & 0.537\,{\scriptsize$\pm$0.117} & $-0.112$ {\scriptsize[$-0.146$, $-0.077$]} & 0.393 {\scriptsize[0.342, 0.442]} \\
500 sensors & 0.310 & 0.141\,{\scriptsize$\pm$0.002} & 0.143 & 0.623 & 0.515\,{\scriptsize$\pm$0.113} & $-0.108$ {\scriptsize[$-0.140$, $-0.076$]} & 0.374 {\scriptsize[0.322, 0.424]} \\
25\% hidden & 0.295 & 0.134\,{\scriptsize$\pm$0.003} & 0.137 & 0.631 & 0.573\,{\scriptsize$\pm$0.121} & $-0.058$ {\scriptsize[$-0.100$, $-0.014$]} & 0.440 {\scriptsize[0.370, 0.511]} \\
60\% hidden & 0.293 & 0.127\,{\scriptsize$\pm$0.002} & 0.126 & 0.651 & 0.577\,{\scriptsize$\pm$0.122} & $-0.074$ {\scriptsize[$-0.114$, $-0.032$]} & 0.450 {\scriptsize[0.389, 0.512]} \\
90\% hidden & 0.302 & 0.153\,{\scriptsize$\pm$0.003} & 0.138 & 0.702 & 0.627\,{\scriptsize$\pm$0.141} & $-0.076$ {\scriptsize[$-0.120$, $-0.027$]} & 0.474 {\scriptsize[0.399, 0.550]} \\
\bottomrule
\end{tabular}
\end{table}

\begin{table}[h]
\caption{Angles held out from training and development. Training uses $11,13,15,17,19^\circ$; testing uses
$12,14,16,18^\circ$ on 16 unseen wing geometries (64 cases). Hidden-region nRMSE and gap closure $G$ computed from
casewise normalised MSE. Intervals group the four conditions of each physical wing.}
\label{tab:held-angle}
\centering
\small
\begin{tabular}{@{}rcccc@{}}
\toprule
Sensors $K$ & Geometry only & Measurements & True angle & $G$ [95\% interval] \\
\midrule
5 & 0.2832 & 0.1947 & 0.1333 & 63.1\% [41.3, 83.0\%] \\
20 & 0.2834 & 0.1550 & 0.1334 & 91.7\% [86.6, 98.6\%] \\
100 & 0.2846 & 0.1464 & 0.1343 & 96.4\% [90.3, 104.4\%] \\
500 & 0.2908 & 0.1433 & 0.1384 & 99.8\% [94.5, 107.3\%] \\
\bottomrule
\end{tabular}
\end{table}

\section{Implementation Details}
\label{app:impl}

\paragraph{Architecture.} Positions enter through Fourier features with eight bands and through 3D rotary embeddings in
every attention layer. The model has width 192 and four heads, with one cross-attention and four self-attention blocks
in the encoder and two decoder blocks, \nParams{} parameters in all.

\paragraph{Field training.} AdamW, one-cycle decay with 10\% warm-up, clipping at 1.0, batch 8, 40 epochs and 2,048
area-sampled queries per example. The matched car completion and decoder studies train all parameters at $10^{-3}$.
The nine-angle DoubleDelta models use $10^{-4}$ for the backbone and $10^{-3}$ for new field parameters.
Normalisation uses training statistics only. Checkpoints minimise development completion error. The held-out-angle
split contains 1,200 training cases, 80 development cases and 64 test cases.

\begin{figure}[H]
\centering
\input{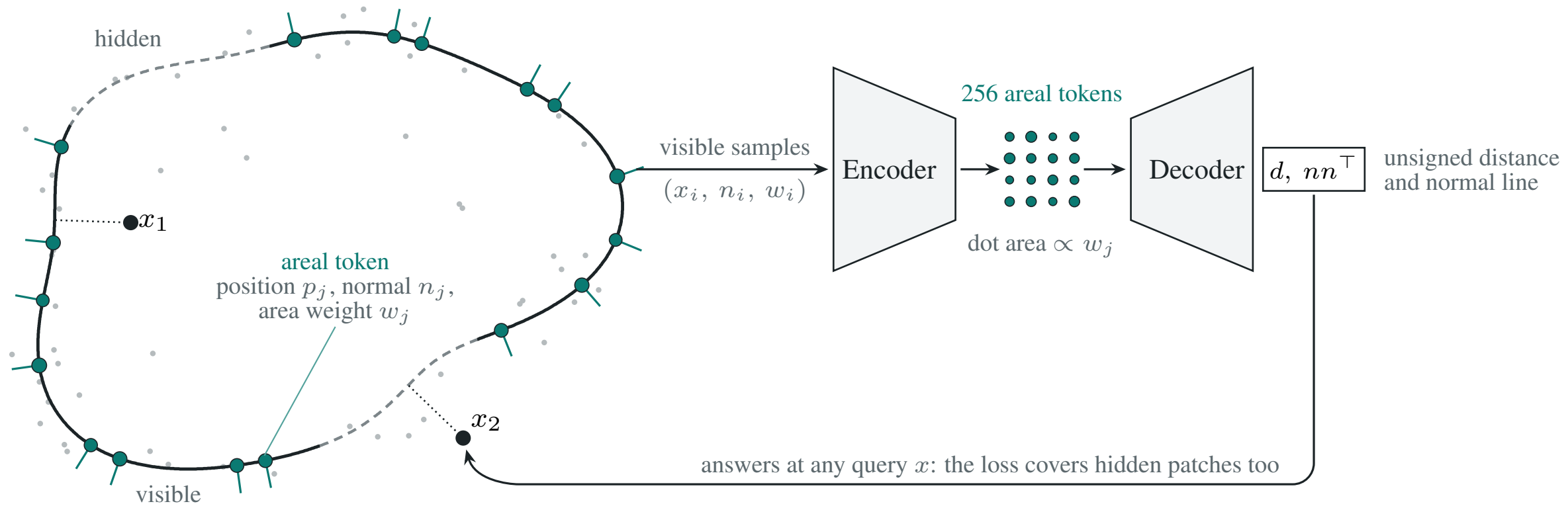}
\caption{\textbf{Masked pretraining into areal tokens.} The encoder reads only the visible surface and writes 256 areal
tokens pinned to it, each with a position, normal and area weight.
The decoder answers unsigned distance and normal at any query point, including $x_2$ over a hidden patch, so the loss
covers the whole surface. Fine-tuning swaps the target: the same encoder reads observed field values, and the decoder
completes the field at any query point (\cref{sec:fieldmode}).}
\label{fig:pretrain}
\end{figure}

\paragraph{Geometry pretraining.} The production run uses 4,096 encoder points, 1,024 decoder queries, 60\% area
masking, 70\% near-surface queries, SO(3) rotations, normal sign flips with probability 0.5, AdamW (learning rate $3\times10^{-4}$, weight decay 0.05, cosine decay)
and bfloat16, at a global batch of 64 on six B200 GPUs. Of the examples, 10\% are shown whole, with a flag that tells the decoder so. The distance target is
in units of the visible surface's area-weighted RMS radius, with a Huber loss of threshold 0.05. The normal loss is the
squared Frobenius distance to the target projector, $2\sin^2\vartheta$ for a unit projector at angle $\vartheta$; under
it the normal output estimates the posterior mean of $\nnT$, whose top eigenvector is a Bayes estimate of the normal
line under the same loss. The corpus is the Sketchfab subset of Objaverse-XL, with 1,844 validation and 1,825 test
shapes held out; at the selected checkpoint, with 60\% of each surface hidden, the validation shapes are reconstructed
with a distance nRMSE of 0.098 and a normal-line error of $22.0^\circ$.
It runs 15 epochs from seed 17; the epoch-11 checkpoint (10,626 updates) minimises development normal error.
The controlled component studies use 8,192 training objects, 128 development and 128 test objects, four epochs
(2,048 updates), seeds 0, 1 and 2, 256 tokens and at most 4,096 input points. They retain SO(3) augmentation and use
the learned heads without analytic surfel or symmetry readouts; input-normal sign augmentation is disabled in every
arm. Production and controlled-study checkpoints are distinct.

\paragraph{Decoder transfer.} All 380-car fits start from geometry-only checkpoints and exclude the 40 evaluation
cars from training; evaluation observes no pressure and scores every surface point. Models use a common
scalar-head initialisation and identical sampled queries, labels and updates
for seeds 0, 1 and 2. The fresh-decoder controls share decoder initialisation. Resetting the \areals{} decoder leaves
all 102 non-decoder state tensors and the scalar head identical. Point-MAE reads XYZ and has 21.825M active backbone
parameters, versus 2.274M for the \areals{} backbone, which also reads normals; a projection maps its 384 channels to
the common width 192. Both encoders produce 256 tokens from at most 4,096 points. Development uses 8,192 fixed queries
per car, while final scores cover every surface point. Training uses BF16 and evaluation FP32. We use 10,000 paired
car-bootstrap draws, retaining all seeds per sampled car, with Bonferroni-adjusted (99.17\%) intervals for the six contrasts among the four pathways. Keeping the pretrained decoder against resetting it gives $[-0.0360,-0.0263]$, and against Point-MAE $[-0.0186,-0.0128]$.

\begin{table}[H]
\caption{The transfer runs through the query decoder: with fresh decoders Point-MAE's larger encoder is better, and
the pretrained decoder reverses this. Full-surface pressure nRMSE ($\downarrow$) on 40 withheld cars, three seeds,
identical field examples, queries and updates; \areals{} rows start from the four-epoch checkpoints
(\cref{tab:ledger}).}
\label{tab:decoder}
\centering
\small
\setlength{\tabcolsep}{4pt}
\begin{tabular}{@{}lc@{}}
\toprule
Transfer pathway & nRMSE \\
\midrule
Point-MAE encoder (21.8M) + fresh query decoder & 0.2556 \\
\areals{} encoder (2.3M) + fresh query decoder & 0.2708 \\
\areals{} encoder + pretrained query decoder & \textbf{0.2399} \\
\bottomrule
\end{tabular}
\end{table}

\paragraph{Reserved-car completion.} The 420/16/48 training/development/test split, checkpoint hashes and protocol
are fixed before fetching test pressures. We average the 25 settings equally within each car, then across cars and
three seeds; no-observation scores are separate. Bootstrap draws retain all settings and seeds within each car.

\paragraph{Frozen surrogate.} The public AeroJEPA tutorial configuration trains for 20 epochs (1,060 updates), with
training-only normalisation and final weights. The geometry prior is frozen. The 48-car replication uses three
downstream seeds and the same 2,048 fixed queries per car. The larger-budget experiments use the token and point
counts in \cref{tab:surrogate}. These frozen-feature experiments use the earlier 61,822-mesh geometry checkpoint
selected at epoch 14; field completion uses the production epoch-11 checkpoint described above.

\paragraph{GABI.} The model, a six-layer graph convolutional network of width 100 with latent dimension 100, its loss
and its normalisation come from the authors' released code. Each 32,000-point geometry gets a symmetrised graph
joining every point to its eight nearest vertices, measured by shortest distance along the edges of the original mesh;
this is the graph reported in the main text. An ablation measures nearness by straight-line distance instead, which
can join the upper and lower surfaces of a thin wing. Each graph is trained from scratch for three seeds, with paired
initial weights and example orders, for 455 epochs (10,010 updates at batch 100) with AdamW, keeping the checkpoint with
the lowest development reconstruction plus MMD loss. We draw 50,000 latent samples, decode them once per geometry and
share that pool across its nine angle cases. ABC acceptance is recomputed for every observation set, with 100
accepted samples and observation noise 0.01 in normalised units. No-observation prediction averages the prior pool.
The along-mesh graph lowers GABI's error at every observation setting, by 12.7\% at 20 sensors, and its force error
by 69\% there (\cref{tab:gabi-full}).

\section{Spatial Interface and Normal-Head Conditioning}
\label{app:architecture}

\paragraph{Normal lines.} In the four-epoch geometry study, all heads are scored by unoriented normal angle on
128 held-out objects. Across three seeds, the projector target gives $25.34^\circ$, against $28.69^\circ$ for an
oriented-vector target and $39.17^\circ$ for an unoriented-vector target. The projector reconstructs normal lines
directly, matching the sign ambiguity of the raw mesh.

\paragraph{A coordinate-aware interface.} On the 48 reserved cars, with no observed pressure, the
surface-anchored, rotary-positioned configuration gives full-surface nRMSE 0.2469 against 0.3898 for the unanchored configuration without rotary positions, a 36.7\%
reduction. This is one of 21 pairwise contrasts among seven architecture variants; its car-bootstrap interval, adjusted over
all 21, is 35.0--38.1\%. These are three
upstream/head initialisations with a shared downstream sampling seed. The comparison measures the joint interface
of anchoring and rotary position encoding.

\paragraph{Isotropic projector initialisation.} Zero weights and bias $(1/3,1/3,1/3,0,0,0)$ start the symmetric
normal head at $I/3$ (diagonal entries first). On one fixed model and batch, changing only this head takes the encoder
normal-loss gradient norm through $155.0\to0\to155.0$ as the original head is replaced and restored.
At $I/3$ the head gradient is 2.17; distance predictions remain identical. Across the matched four-epoch runs,
distance MSE falls from 0.01167 to 0.00854 (26.8\%) and normal error from $25.34^\circ$ to $24.87^\circ$.
Pressure nRMSE on the 40-car study falls from 0.2564 to 0.2399 (6.4\%), improving in all three seeds, with adjusted
paired interval $[-0.01960,-0.01378]$.

\section{Finite-Token and Density Tests}
\label{app:stability}

\begin{figure}[H]
\centering
\includegraphics[width=0.92\textwidth]{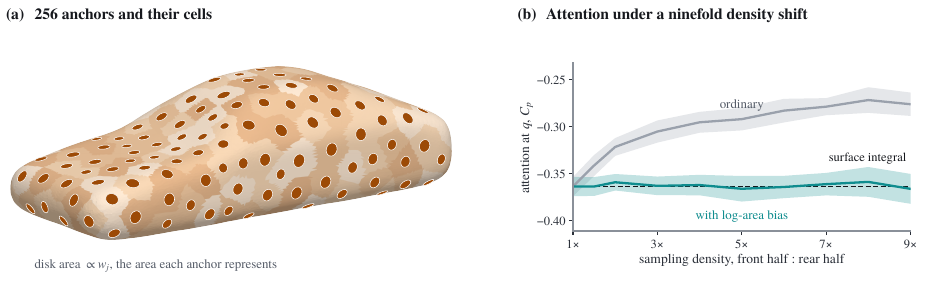}
\caption{\textbf{Anchors as a quadrature}, a structural analogy rather than a theorem. (a)~256 farthest-point anchors on
the car of \cref{fig:hero}, with disks sized by the area $w_j$ each represents. (b)~Attention at a query above the roof,
a kernel average of $C_p$ over 4,096 samples, as the front half is sampled up to nine times more densely; bands are one
standard deviation over 32 draws. Ordinary attention drifts by 24\%; with the log area weight in its logits, attention
stays on the surface integral (dashed).}
\label{fig:quadrature}
\end{figure}

\paragraph{Density shift.} On 32 meshes, three geometry-pretraining seeds and three density draws per seed, one
geometric half is sampled ninefold more densely than the other. Inverse-proposal weights represent the same area
measure. With anchors and frame fixed, normalised output drift is 0.002535 for log-area attention and 0.013706 for
ordinary attention. The paired drift difference is $-0.011171$, with object-grouped 95\% interval
$[-0.012692,-0.009690]$. All 1,344 saved prediction records are independently checked. This experiment evaluates
the quadrature component under a changed sampling density. On the car of \cref{fig:hero}, the same shift moves ordinary attention at one query by 24\%, while log-area attention
stays within 0.7\% of the surface integral (\cref{fig:quadrature}).

\begin{table}[h]
\caption{Finite-token prediction sensitivity on 32 raw meshes using a fixed production checkpoint. Values are mean
RMS changes in unsigned-distance prediction, divided by the object's RMS radius. Coupled coplanar subdivision keeps
the sampled surface points fixed, so the first two rows are invariant by construction and serve as sanity checks. The resampling control fixes anchors, frame and surfels.}
\label{tab:finite-token}
\centering
\small
\begin{tabular}{@{}lrr@{}}
\toprule
Perturbation & 256 tokens & 1,024 tokens \\
\midrule
Point order & 2.27e-08 & 1.58e-08 \\
Coupled coplanar subdivision & 6.60e-09 & 5.20e-09 \\
Independent surface resampling & 0.01316 & 0.00921 \\
Resampling, fixed anchors/frame/surfels & 0.00467 & 0.00217 \\
\bottomrule
\end{tabular}
\end{table}

\paragraph{Attention splitting.} In a hand-checked attention fixture, splitting weighted keys while preserving
their total area changes log-area attention by RMS $1.49\times10^{-8}$, against 0.03192 for ordinary attention.
The fixed-checkpoint experiment saves 448 prediction cases across the two token settings. Independent recomputation
agrees on output change to $2.3\times10^{-9}$ and distance NMSE to $4.4\times10^{-19}$.

\paragraph{Averaging resampled views.} On the same 32 meshes with 60\% of the surface hidden, averaging four
independently resampled predictions lowers all-query normal error from $19.06^\circ$ to $18.53^\circ$ and hidden-query
error from $24.72^\circ$ to $24.26^\circ$. Distances are averaged directly; normals are combined through the mean
projector and its leading eigenvector. This uses the frozen production checkpoint throughout.

\section{Benchmark Controls}
\label{app:controls}

\paragraph{Hidden-value perturbation.} We replace hidden pressures in a scratch copy of the input files and rerun
the trained predictor. The allowed change is $10^{-3}$ of the field spread. The 48-car study passes 264 checks;
the 40-car decoder study passes 96; the held-out-angle study passes 36. GABI passes all six recorded checks, with
maximum changes below $8\times10^{-6}$ of the field spread, and the pilot of the three-seed GABI runs repeats this check
and the parity of our ABC step with the authors'.

\paragraph{Measurement response.} Increasing observed pressure by one field spread changes GABI's hidden prediction
by approximately 0.10 spreads at 20 sensors. The measurement model's condition input is disabled. In the held-out-angle
execution, blocked condition and search artifacts are never read, the condition-construction function is never
called, and one encoder invocation yields 32,000 output values through chunked queries.

\paragraph{Independent recomputation.} Scores from saved predictions, targets and area weights agree within
$3.4\times10^{-16}$ on reserved cars, $2.0\times10^{-16}$ on decoder transfer and $1.2\times10^{-12}$ on held-out
angles; for GABI, all 864 test predictions of the six runs at 20 sensors reproduce the recorded scores within
$3.4\times10^{-14}$, and for \areals{}, all 6,912 predictions of the three seeds, over eight observation settings on
the development and test splits, within $4\times10^{-14}$. We check split identities and record checkpoint hashes before and after inference.

\section{Symmetry Readouts}
\label{app:priors}

\begin{figure}[H]
\centering
\input{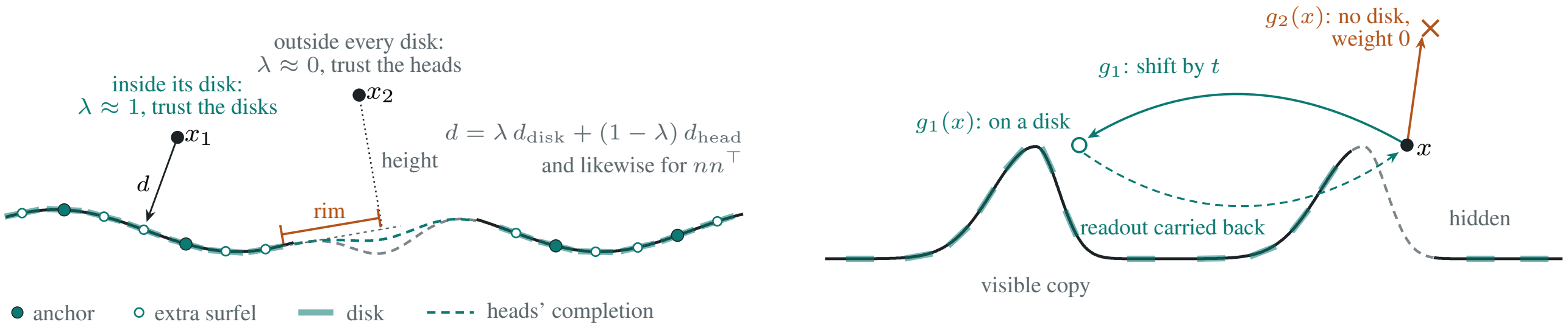}
\caption{\textbf{Surfel and symmetry readouts.} (a) A query over visible surface ($x_1$) projects inside its nearest disk, so
the gate trusts the disks ($\lambda\approx1$); over a hole ($x_2$) it projects outside every disk, and the gate, told
how far outside (rim) and how high above (height), hands over to the learned heads. (b) $g_1$ shifts one repeated part
onto the other, so for a query $x$ over the hole the surfel readout at $g_1(x)$ is carried back to $x$; $g_2$ holds
elsewhere, but its image lands on no disk, so it gets no weight.}
\label{fig:priors}
\end{figure}

\begin{table}[H]
\caption{Symmetry maps repair hidden copies of visible parts and leave visible geometry untouched. Normal error in
degrees ($\downarrow$) on 41 held-out meshes, one seed; the gate is trained on 170 meshes. All is pooled over points;
Hidden is averaged over meshes.}
\label{tab:geometry}
\centering
\small
\setlength{\tabcolsep}{4.5pt}
\begin{tabular}{@{}lcccc@{}}
\toprule
 & Full & \multicolumn{3}{c}{60\% of the surface hidden} \\
\cmidrule(l){3-5}
Prior & input & All & Hidden & Repeated parts \\
\midrule
Surfels, 8 per token & 5.7 & 12.4 & 17.6 & 10.9 \\
+ symmetry maps & 5.7 & \textbf{11.4} & \textbf{15.7} & \textbf{6.9} \\
\bottomrule
\end{tabular}
\end{table}

\paragraph{Symmetry makes hidden parts observable.} On hidden points that have a visible copy, symmetry maps cut normal
error from $10.9^\circ$ to $6.9^\circ$, and across all hidden points from $17.6^\circ$ to $15.7^\circ$, improving 32 of
41 held-out meshes while leaving visible geometry untouched (\cref{tab:geometry}).

On hidden points with a visible copy, the gate gives the symmetry maps 61\% of the weight, and removing the maps at
inference restores the baseline. With symmetries detected on the complete shapes, 63\% of hidden points map
onto visible input, so the same mechanism reaches further as detection improves.

\section{Distance Tubes}
\label{app:tubes}

\paragraph{Protocol.} The production checkpoint reads the first eight test-split meshes in identifier order and three
analytic surfaces: a unit sphere, a torus with radii 1 and 0.35, and two spheres of radius 0.55 whose centres are 1.3
apart. Each shape gives 16,384 area-weighted surface samples, at most 4,096 encoder points and 2,048 surfels, either
whole or with 60\% of its area hidden as 32 patches. Distances are in units of the complete shape's area-weighted RMS radius and
are evaluated on a $97^3$ grid with 0.6 of padding, against the exact distance to the mesh triangles or the analytic
surface. At nine radii from 0.025 to 0.4, each tube $\{\hat d\le r\}$ gives three quantities: the area estimate
$\mathrm{Vol}\{\hat d\le r\}/(2r)$, which for a closed $C^2$ surface below its reach equals
$A(S)+\tfrac{2\pi}{3}\chi(S)\,r^2$ \citep{weyl1939tubes}; the Betti numbers over $\mathbb{F}_2$ of the union of closed
voxels; and the shortest-path lengths between six farthest-point landmarks through the 26-connected occupied voxels,
which converge to geodesic distances as the tube thins \citep{memoli2005distance}. Each is compared with the exact
distance's tube at the same radius. Persistence diagrams of the sublevel filtrations up to $r=0.4$ are compared by
bottleneck distance, the quantity \cref{lem:primitive} bounds.

\paragraph{Complete inputs.} A complete input returns the distance to the 2,048 disks kept with the tokens, each through a surfel,
perpendicular to its normal and of radius half the median surfel spacing. On the analytic shapes this distance exceeds
the exact one by at most 0.07, in the gaps between disks, and never falls below it by more than 0.001; from $r=0.1$,
just above those gaps, every tube has the exact Betti numbers, area within 1.4\% and path lengths within 2.4\% of the
exact tube (\cref{tab:tubes}). On the eight held-out meshes the disk distance never falls below the exact one by more
than 0.045 anywhere on the grid. Tube volume matches the exact tube's within 7.1\% from $r=0.1$ and within 3.5\% from
$r=0.28$, every landmark pair stays connected from $r=0.14$, and path lengths agree within 3.6\% from $r=0.2$. On all
eleven shapes the persistence diagrams lie within bottleneck distance 0.063, 0.051 and 0.101 of the exact ones in
dimensions 0, 1 and 2, so every component, loop and cavity that persists beyond 0.21 RMS radii is reproduced with birth
and death within 0.11, and no added feature persists beyond 0.21.

\begin{table}[H]
\caption{Complete inputs: tubes of the decoder's distance against the exact distance's tubes at the same radius,
maxima over shapes. Analytic shapes are the sphere, torus and two spheres; meshes are the eight held-out meshes.}
\label{tab:tubes}
\centering
\small
\begin{tabular}{@{}lcccc@{}}
\toprule
 & \multicolumn{3}{c}{Analytic shapes} & Meshes \\
\cmidrule(lr){2-4}\cmidrule(l){5-5}
Radius $r$ & Exact Betti numbers & Area error (\%) & Path error (\%) & Area error (\%) \\
\midrule
0.10 & 3 of 3 & 1.3 & 2.4 & 7.1 \\
0.14 & 3 of 3 & 0.8 & 0.4 & 5.6 \\
0.20 & 3 of 3 & 0.3 & 0.0 & 4.9 \\
0.28 & 3 of 3 & 0.1 & 0.1 & 3.5 \\
0.40 & 3 of 3 & 0.1 & 0.5 & 2.9 \\
\bottomrule
\end{tabular}
\end{table}

\paragraph{Partial inputs.} With 60\% of the area hidden, the decoder's distance covers the hidden surface
(\cref{tab:tubes-masked}). It falls below 0.05 RMS radii over 79--98\% of the hidden area of each held-out mesh, over at
least 92\% on six of the eight, and below 0.1 over 83--99.5\%, against 8--16\% and 19--37\% for disks built from the
visible surface; on the analytic shapes it covers 98--100\% of the hidden area at 0.05. Its median over the hidden
surface is 0.018--0.025, against 0.13--0.24 for the visible disks, and it stays below 0.05 over at least 99.8\% of the
visible area. On the held-out meshes its persistence diagrams lie within bottleneck distance 0.053, 0.090 and 0.119 of
the exact ones in dimensions 0, 1 and 2, against 0.200, 0.110 and 0.190 for the visible disks: completion brings the
distance closer to the exact one in topology as well as in position.

\begin{table}[H]
\caption{Partial inputs, 60\% of the area hidden: share of the hidden area (\%, $\uparrow$) over which the distance
falls below 0.05 and 0.1 RMS radii, for \areals{} and for disks built from the visible surface. Meshes 1--8 are the
held-out meshes in identifier order.}
\label{tab:tubes-masked}
\centering
\small
\begin{tabular}{@{}lcccc@{}}
\toprule
 & \multicolumn{2}{c}{Below 0.05} & \multicolumn{2}{c}{Below 0.1} \\
\cmidrule(lr){2-3}\cmidrule(l){4-5}
Shape & \areals{} & Visible disks & \areals{} & Visible disks \\
\midrule
Mesh 1 & \textbf{96.1} & 9.1 & \textbf{99.5} & 23.9 \\
Mesh 2 & \textbf{92.2} & 11.5 & \textbf{95.2} & 30.3 \\
Mesh 3 & \textbf{78.9} & 9.2 & \textbf{83.1} & 23.4 \\
Mesh 4 & \textbf{97.6} & 8.9 & \textbf{99.0} & 20.3 \\
Mesh 5 & \textbf{93.7} & 16.4 & \textbf{97.6} & 37.2 \\
Mesh 6 & \textbf{97.1} & 8.5 & \textbf{99.4} & 23.1 \\
Mesh 7 & \textbf{86.3} & 12.5 & \textbf{93.9} & 31.9 \\
Mesh 8 & \textbf{94.7} & 7.8 & \textbf{97.1} & 18.9 \\
\midrule
Sphere & \textbf{97.6} & 10.8 & \textbf{100.0} & 23.4 \\
Torus & \textbf{100.0} & 12.1 & \textbf{100.0} & 27.1 \\
Two spheres & \textbf{100.0} & 11.9 & \textbf{100.0} & 25.9 \\
\bottomrule
\end{tabular}
\end{table}

\section{Attention Range}
\label{app:attention}

\begin{figure}[H]
\centering
\input{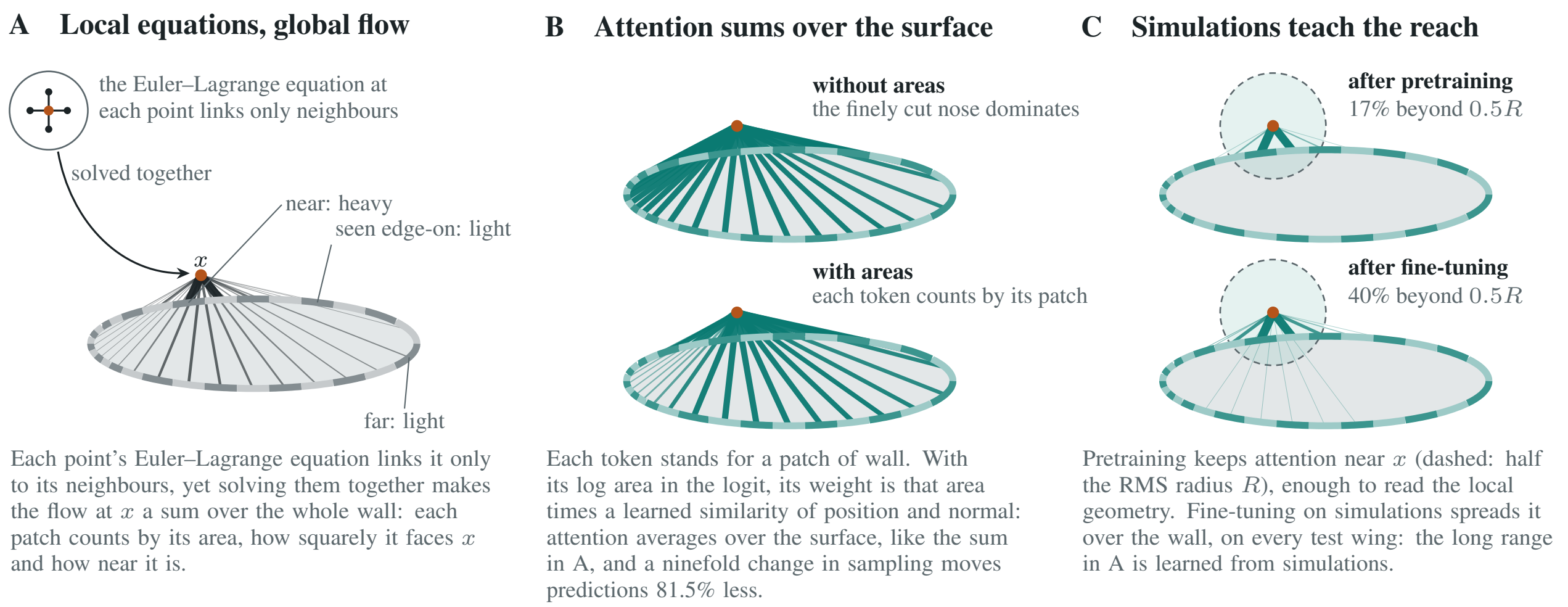}
\caption{\textbf{The Euler--Lagrange equation is local, its solution is global, and simulations teach the reach.} Every panel uses the same
2D wing, patches and query $x$. (A) Line widths are the double-layer kernel times patch area for attached flow, which is
the angle each patch subtends at $x$. (B) Widths use one illustrative similarity, without and with the log-area bias of
\cref{eq:attn}. (C) Widths are fitted so that the share beyond half the RMS radius $R$ equals the measured share of
decoder attention on the 16 DoubleDelta test wings.}
\label{fig:attention}
\end{figure}

\paragraph{Protocol.} Three checkpoints read the 144 DoubleDelta test cases (16 wings, nine angles) with the whole
surface and no sensors: the production geometry checkpoint in the field architecture with its new input columns at
zero, the sensor-conditioned \areals{} model fine-tuned from it, and the same transformer trained from scratch. For
2,048 area-sampled surface queries per case, the attention weights of both decoder cross-attention layers and all four
heads, including the log-area bias, give the attention-weighted distance from query to token anchor in units of the
surface's area-weighted RMS radius $R$. The reference is attention in proportion to area, which every query pays when
all logits are zero. Self-attention among the encoder's tokens is measured the same way. Each wing contributes the mean
over its cases; intervals are Student-t over the 16 wings, with exact sign-flip tests.

\paragraph{Fine-tuning adds long-range attention.} The pretrained decoder places 52\% of its attention within $0.25R$
of the query, where attention in proportion to area places 3\%, and 17\% beyond $0.5R$, against 88\%. Fine-tuning
raises the share beyond $0.5R$ to 40\%, an increase of 0.230 [0.220, 0.239] that holds on every wing (exact one-sided
$p=2^{-16}$), and its mean attention distance rises from 0.25 to 0.41 of the area-proportional one (\cref{tab:attention}).
The last decoder layer carries most of the change, from 12\% to 46\% beyond $0.5R$. The encoder's token self-attention
follows the same pattern, from 47\% to 59\% beyond $0.5R$ (+0.126 [0.123, 0.129], every wing). Trained from scratch on
the same simulations, the transformer spreads its attention almost in proportion to area: 72\% of the decoder's and
86\% of the encoder's attention lies beyond $0.5R$, so the local structure of \areals{} attention comes from geometry
pretraining.

\begin{table}[H]
\caption{Attention range on the 16 DoubleDelta test wings, no sensors. Mean attention distance relative to attention in
proportion to area, with 95\% intervals over wings, and the share of attention (\%) by distance from the query in
units of the RMS radius $R$.}
\label{tab:attention}
\centering
\small
\begin{tabular}{@{}llcccc@{}}
\toprule
 & Checkpoint & Distance ratio & Within $0.25R$ & Beyond $0.5R$ & Beyond $R$ \\
\midrule
Decoder & Pretrained & 0.254 {\scriptsize[0.251, 0.256]} & 51.5 & 17.0 & 3.5 \\
 & Fine-tuned & 0.406 {\scriptsize[0.400, 0.412]} & 27.5 & 40.0 & 11.0 \\
 & From scratch & 0.786 {\scriptsize[0.775, 0.796]} & 10.6 & 72.4 & 42.3 \\
 & In proportion to area & 1 & 3.2 & 88.5 & 63.6 \\
\midrule
Encoder & Pretrained & 0.524 {\scriptsize[0.522, 0.527]} & 23.4 & 46.7 & 21.2 \\
 & Fine-tuned & 0.648 {\scriptsize[0.642, 0.654]} & 18.9 & 59.3 & 31.9 \\
 & From scratch & 0.983 {\scriptsize[0.982, 0.985]} & 4.1 & 86.5 & 60.6 \\
 & In proportion to area & 1 & 3.3 & 88.2 & 62.5 \\
\bottomrule
\end{tabular}
\end{table}

\paragraph{Field error by region.} With neither sensors nor angle, each model returns one field per wing, so its
squared error splits exactly into its distance from the mean of the wing's nine angle-specific fields and the variation
of those fields about their mean, which no such model can predict. Against the mean field, fine-tuned \areals{}
removes 58.8\% [53.7, 63.8] of the squared error of the model trained from scratch over the whole surface,
55.6\% [45.3, 66.0] over the most curved third of each surface at ordinary pressures and 61.4\% [51.2, 71.6] over the
flattest third at extreme pressures, on every wing. Regions are terciles by area of the
shape-operator norm, fitted from the projector field over each point's nearest neighbours through the identity of
\cref{lem:primitive}, and of the mean field's distance from its median in units of its spread.

\section{Frozen Areal Tokens as a Surrogate's Geometry Input}
\label{app:frozen}

\begin{table}[H]
\caption{Frozen areal tokens as a surrogate's geometry input. Mean per-car relative $L_2$ error ($\downarrow$) on
2,048 fixed pressure queries, 16 DrivAerML validation cars. The first two blocks are matched pairs, with the better
entry in bold; the last block lists other budgets, which are not matched. $^\dagger$Interpolated from 64 point groups.}
\label{tab:surrogate}
\centering
\small
\begin{tabular}{@{}lccc@{}}
\toprule
Geometry prior & Tokens & Points & Rel.\ $L_2$ \\
\midrule
\areals{} & 256 & $\le$4,096 & \textbf{0.3283} \\
AeroJEPA tutorial & 256 & $\le$4,096 & 0.3338 \\
\midrule
\areals{} & 1,024 & $\le$4,096 & \textbf{0.3244} \\
Utonia \citep{utonia2026} & 1,024 & $\le$4,096 & 0.3294 \\
\midrule
Point-MAE \citep{pang2022pointmae} & 1,024$^\dagger$ & 1,024 & 0.3353 \\
\areals{} & 1,024 & 65,536 & 0.3198 \\
\areals{} & 32,768 & 65,536 & 0.3073 \\
\bottomrule
\end{tabular}
\end{table}

Areal tokens also transfer without fine-tuning. As the frozen geometry input of the AeroJEPA tutorial surrogate
\citep{aerojepa}, trained for 20 epochs on the same 420 cars, they lower error by 1.65\% against its own encoder at
matched 256-token and 4,096-point budgets (\cref{tab:surrogate}). On the 48 reserved cars, the matched three-seed
comparison gives 0.32885 against 0.33193, a 0.9\% reduction with paired interval $[-0.00571,-0.00034]$. Reading the
same frozen checkpoint with more tokens and points lowers its error further, to 0.3073 at 32,768 tokens and 65,536
points. These experiments use the earlier \nMeshes{}-mesh geometry checkpoint selected
at epoch 14 (\cref{app:impl}). A companion study \ifblind\citep{anon2026reading}\else\citep{chlon2026reading}\fi{}
reads the same tokens through a frozen vision--language model.

\section{Matched Comparison of Pretraining Targets}
\label{app:targets}

\paragraph{Cohort and priors.} At a snapshot of the preparation run, 5,287 objects had been processed, 2,123 had
passed patch preparation and 4,529 the query preparation. The cohort is every object that passed both: 1,668
training, 85 validation and 92 test meshes in the original Objaverse-XL split and order, with no replacements. It is a
preparation-success subset, not a random sample. Each prior trains from fresh weights and a fresh optimiser on
complete, unmasked inputs for 10,626 updates at a global batch of 64 objects (680,064 presentations), with seed 17 on
four B200 GPUs, a 200-update warm-up, cosine decay to 5\% of the peak rate and clipping at 1, and otherwise keeps its
own recipe (\cref{tab:target-recipes}); data and updates are matched, FLOPs are not. The \areals{} arm uses this
paper's implementation with targets $d$ and $\nnT$ and computes its forward pass in bfloat16. The Hessian arm is a
separately implemented \areals{}-sized query network that adds the smoothed Hessian $H_r$ of $\eta_S$ at several radii,
estimated without bias from nearest-point queries through the divergence theorem; near a smooth wall $H_r$ is a
smoothed $\nnT+tW(I+tW)^{-1}$, so it carries the shape operator. The patch arm predicts, for eight area-sampled patches
per object, the coefficients of a polynomial chart in normal coordinates, from which $A=|\det J|\,g^{-1}$ follows
analytically, and minimises the Frobenius error of $A$ under an augmented-Lagrangian bound of 0.01 bounding-box
diagonals on the RMS position error; its encoder runs in FP32 and its geometry in FP64, and it does not infer distances
to the union of its patches.

\begin{table}[h]
\caption{The three priors: recipe and geometry on the 92 test meshes, as per-object means. Patch geometry is scored
against fitted reference charts on eight sampled patches per object and the other two against mesh queries, so the
geometry entries do not rank the representations.}
\label{tab:target-recipes}
\centering
\small
\setlength{\tabcolsep}{4pt}
\begin{tabular}{@{}lr>{\raggedright\arraybackslash}p{2.6cm}cc>{\raggedright\arraybackslash}p{2.4cm}c>{\raggedright\arraybackslash}p{3.0cm}@{}}
\toprule
Arm & Parameters & Targets & Peak rate & Weight decay & Distance or position & Normal line & Other \\
\midrule
\areals{} & 3,215,503 & $d$, $\nnT$ & $3\times10^{-4}$ & 0.05 & distance nRMSE 0.0089 & $11.0^\circ$ & --- \\
Hessian & 3,217,948 & $d$, $\nnT$, $H_r$ & $3\times10^{-4}$ & 0.05 & distance nRMSE 0.0183 & $15.0^\circ$ & $H$ error 0.495 (label floor 0.007) \\
Patch & 2,268,298 & $A$, position-constrained & $10^{-4}$ & 0 & position RMS 0.0013 & $1.6^\circ$ & relative $A$ RMS 0.019 \\
\bottomrule
\end{tabular}
\end{table}

\paragraph{Fine-tuning and testing.} WindsorML's 274 training, 35 validation and 35 test cars keep the nested budgets of
\cref{fig:efficiency}: 18, 35, 69, 137 and 274 cars. Every model predicts pressure from complete geometry with no
sensors and trains all weights for 8,000 updates of eight cars and 2,048 queries per car, with AdamW in FP32, constant
rates, no weight decay, a rate of $2\times10^{-4}$ for new parameters and validation every 250 updates. The encoder's
learning rate is chosen from $\{2\times10^{-5},2\times10^{-4},10^{-3}\}$ on seed 0 separately for each arm, initialisation and
budget, and seeds 1 and 2 repeat the chosen rate. All 150 fits completed. The 90 selected runs, three seeds for each
arm, initialisation and budget, were fixed on validation alone and locked by hash before any of them was scored on the
test cars. The test split had been scored once before, by the patch pilot below; no selection used that score. The
score is the mean over cars of the area-weighted RMSE of $C_p$ divided by its area-weighted standard deviation.

\paragraph{Results.} \Cref{tab:targets-full} gives every comparison on the test cars, with the validation scores
beside them. The \areals{} and Hessian gains exclude zero at every budget, and the patch intervals exclude zero at the
three largest budgets, where pretraining raises the error. Pretrained \areals{} has lower error than the pretrained
Hessian arm at every budget, by 0.0238, 0.0184, 0.0127, 0.0008 and 0.0025 nRMSE from 18 to 274 cars; the paired 95\%
intervals of the difference exclude zero at 18, 35 and 69 cars and include it at 137 and 274 ($[-0.0104,0.0103]$ and
$[-0.0207,0.0135]$, \areals{} minus Hessian). To reach a test nRMSE of 0.25, the \areals{} and Hessian priors need 35
cars, every model from scratch 137 and the patch prior 274.

\begin{table}[h]
\caption{Matched pretraining targets, every comparison: test nRMSE ($\downarrow$) over the 35 test cars, means of three
fine-tuning seeds, with validation means for reference. The difference is pretrained minus scratch on test, averaged
over seeds and cars, with a 95\% interval from 5,000 paired bootstrap draws of seeds and cars.}
\label{tab:targets-full}
\centering
\small
\setlength{\tabcolsep}{5pt}
\begin{tabular}{@{}lrcccccc@{}}
\toprule
 & & \multicolumn{5}{c}{Test} & Validation \\
\cmidrule(lr){3-7}\cmidrule(l){8-8}
Arm & Cars & Pretrained & Scratch & Gain & Difference & 95\% interval & Pretrained / scratch \\
\midrule
\areals{} ($d,\nnT$) & 18 & \textbf{0.2773} & 0.4431 & 37.4\% & $-0.1658$ & $[-0.1929,-0.1447]$ & 0.2649 / 0.4413 \\
 & 35 & \textbf{0.2281} & 0.3809 & 40.1\% & $-0.1528$ & $[-0.1817,-0.1266]$ & 0.2114 / 0.3602 \\
 & 69 & \textbf{0.1934} & 0.2649 & 27.0\% & $-0.0715$ & $[-0.0949,-0.0512]$ & 0.1769 / 0.2649 \\
 & 137 & \textbf{0.1778} & 0.2279 & 22.0\% & $-0.0501$ & $[-0.0733,-0.0324]$ & 0.1506 / 0.2037 \\
 & 274 & \textbf{0.1582} & 0.2026 & 21.9\% & $-0.0443$ & $[-0.0661,-0.0247]$ & 0.1410 / 0.1837 \\
\midrule
Hessian ($d,\nnT,H_r$) & 18 & 0.3011 & 0.4753 & 36.7\% & $-0.1742$ & $[-0.2076,-0.1401]$ & 0.2879 / 0.4721 \\
 & 35 & 0.2465 & 0.3882 & 36.5\% & $-0.1417$ & $[-0.1817,-0.1077]$ & 0.2331 / 0.3727 \\
 & 69 & 0.2061 & 0.3006 & 31.4\% & $-0.0945$ & $[-0.1260,-0.0644]$ & 0.1892 / 0.2950 \\
 & 137 & 0.1786 & 0.2147 & 16.9\% & $-0.0362$ & $[-0.0489,-0.0230]$ & 0.1606 / 0.2079 \\
 & 274 & 0.1607 & 0.1822 & 11.8\% & $-0.0215$ & $[-0.0327,-0.0103]$ & 0.1452 / 0.1735 \\
\midrule
Patch ($A$) & 18 & 0.4575 & 0.4419 & $-3.5$\% & $+0.0155$ & $[-0.0218,+0.0614]$ & 0.4401 / 0.4231 \\
 & 35 & 0.3838 & 0.3789 & $-1.3$\% & $+0.0049$ & $[-0.0456,+0.0521]$ & 0.3748 / 0.3649 \\
 & 69 & 0.3068 & 0.2730 & $-12.4$\% & $+0.0337$ & $[+0.0119,+0.0539]$ & 0.3079 / 0.2735 \\
 & 137 & 0.2568 & 0.2233 & $-15.0$\% & $+0.0335$ & $[+0.0097,+0.0575]$ & 0.2450 / 0.2257 \\
 & 274 & 0.2212 & 0.1956 & $-13.1$\% & $+0.0256$ & $[+0.0077,+0.0474]$ & 0.2174 / 0.1894 \\
\bottomrule
\end{tabular}
\end{table}

\paragraph{Patch pilot.} A first patch prior, pretrained on 385 meshes for 10,626 updates with seed 17, was fine-tuned
on all 274 training cars with one pressure seed, choosing the encoder's learning rate from the same grid on validation for both
initialisations; both chose $2\times10^{-5}$. On the 35 test cars it scored 0.2176 against 0.1917 from scratch, 13.5\%
worse, close to the 13.1\% of the full comparison at 274 cars.

\paragraph{Coefficients inside \areals{}.} The patch prior differs from \areals{} in architecture, input
representation and arithmetic as well as in target. A fourth prior, the coefficient prior, trains the coefficients
inside \areals{}'s own encoder and query decoder. Its loss is the distance and normal-line loss of \areals{} plus 0.1
times an auxiliary chart loss: position and normal line at patch anchors, in units of the patch radius, and the
relative error of $A$, computed analytically from cubic chart coefficients that the shared query decoder predicts. It
is pretrained with seed 17 for 10,626 updates at a global batch of 64 on a larger cohort of 14,304 training, 637
validation and 632 test meshes, in FP32 and with neither masking nor rotation, and its whole encoder and query decoder
transfer to pressure under a fresh scalar head. Fine-tuning follows the protocol above, except that every seed searches
its own encoder learning rate.

Against the same architecture from scratch, the coefficient prior lowers the test error by 18--38\% at every budget,
with every interval excluding zero, and it is 29--40\% more accurate than the patch prior (\cref{tab:coefficient-prior}).
Trained on distance and normal line alone, with pretraining seeds 17 and 29 and fine-tuning seed 0, the same network,
cohort and recipe lowers the error of scratch by 25--41\%, and the two priors differ by at most 0.007 nRMSE, with
intervals over cars that include zero at every budget. Coefficient supervision neither helps nor hurts once the global
distance field is learned; without that field, as in the patch prior, the coefficients do not transfer. The scratch
fits are those of the \areals{} arm above wherever the recipe matches. At four cells, two seeds at 274 cars and one
each at 18 and 35, the per-seed rate search lowers the error from scratch, at 274 cars from 0.203 to 0.192, so these
gains are measured against a slightly stronger baseline.

\begin{table}[h]
\caption{Coefficient targets inside \areals{}: test nRMSE ($\downarrow$) on the 35 test cars. Left: the coefficient
prior against the same architecture from scratch, means of three fine-tuning seeds, with the difference (prior minus
scratch) and its 95\% interval from 5,000 paired bootstrap draws of seeds and cars. Right: fine-tuning seed 0, means
over pretraining seeds (three for the coefficient prior, two for distance and normal line alone), with the difference
(coefficients minus distance and normal line alone) and its 95\% interval from 5,000 bootstrap draws of cars.}
\label{tab:coefficient-prior}
\centering
\small
\setlength{\tabcolsep}{4pt}
\begin{tabular}{@{}rcccccc@{}}
\toprule
 & \multicolumn{3}{c}{Three fine-tuning seeds} & \multicolumn{3}{c}{Fine-tuning seed 0} \\
\cmidrule(lr){2-4}\cmidrule(l){5-7}
Cars & Coefficient prior & Scratch & Difference [95\% interval] & Coefficients & $d,\nnT$ alone & Difference [95\% interval] \\
\midrule
18 & 0.2789 & 0.4359 & $-0.1569$ $[-0.1812,-0.1331]$ & 0.2637 & 0.2648 & $-0.0010$ $[-0.0063,+0.0041]$ \\
35 & 0.2304 & 0.3729 & $-0.1424$ $[-0.1645,-0.1213]$ & 0.2222 & 0.2248 & $-0.0027$ $[-0.0090,+0.0041]$ \\
69 & 0.1896 & 0.2649 & $-0.0753$ $[-0.0963,-0.0572]$ & 0.1874 & 0.1876 & $-0.0002$ $[-0.0080,+0.0062]$ \\
137 & 0.1690 & 0.2279 & $-0.0589$ $[-0.0759,-0.0439]$ & 0.1730 & 0.1688 & $+0.0043$ $[-0.0021,+0.0112]$ \\
274 & 0.1565 & 0.1915 & $-0.0350$ $[-0.0487,-0.0216]$ & 0.1552 & 0.1482 & $+0.0069$ $[-0.0004,+0.0168]$ \\
\bottomrule
\end{tabular}
\end{table}

\paragraph{Pretraining seeds.} The coefficient prior was pretrained twice more, with seeds 29 and 43, and each copy was
fine-tuned with seed 0. Across the three pretraining seeds, the standard deviation of the test error lies between
0.0016 and 0.0068 nRMSE at the five budgets; their means are in \cref{tab:coefficient-prior}.

\paragraph{Scope.} Test scores were computed only after every selection was locked. The three original arms differ in
implementation, input representation, preprocessing and arithmetic as well as in target, so they compare the
pretraining gains of three representations rather than isolating one target; the coefficient prior isolates the
coefficient target within one architecture and recipe, but not the curvature target. The largest encoder learning rate was
chosen in 9 of the 30 cells, six of them from scratch, and the patch arm chose the smallest in eight of its ten, so
some optima may lie outside the grid; 73\% of the 150 fits reach their best validation score at or after update 7,000
of 8,000, so the fixed budget is not a claim of convergence. A checkpoint of the \areals{} prior saved at 6,500 updates
scored 0.2522 at 18 cars on seed 0, against 0.2734 for the final one, so more geometry updates did not monotonically
improve transfer here. These priors are pretrained on 1,668 meshes and the coefficient prior on 14,304, not the
\nMeshes{} of the production checkpoint used elsewhere in the paper.

\section{GeoPT and \areals{}}
\label{app:geopt}

GeoPT \citep{geopt2026} also pretrains on geometry for physics, so it is the comparison readers ask for first. The two
methods differ in what they pretrain, what their networks take as input and what they estimate (\cref{tab:geopt}), and
GeoPT cannot enter the sensor-conditioned comparison of \cref{sec:experiments} without being changed into a different
model.

\begin{table}[h]
\caption{GeoPT and \areals{} side by side; GeoPT entries follow its paper \citep{geopt2026}.}
\label{tab:geopt}
\centering
\small
\setlength{\extrarowheight}{2.5pt}
\begin{tabular}{@{}>{\raggedright\arraybackslash}p{3.0cm}>{\raggedright\arraybackslash}p{6.4cm}>{\raggedright\arraybackslash}p{6.4cm}@{}}
\toprule
 & GeoPT & \areals{} \\
\midrule
Pretraining task & Regress geometric features, such as signed distance, along straight trajectories of points given random velocities, halted at the wall & Complete raw surfaces with 60\% of the area hidden: unsigned distance and normal line at any point \\
Origin of the targets & Designed: synthetic dynamics that obey no conservation law & Derived: in wall coordinates a change of shape is a change of the flow equation's coefficients, fixed to first order by distance and normal line (\cref{lem:primitive,prop:outer}); reconstruction is sufficient and both targets are necessary (\cref{prop:sufficiency}) \\
Network inputs & Geometry and a velocity field, in pretraining and in fine-tuning & Geometry; once fine-tuned, also field values and their presence flags \\
Geometry-only model & None & The pretrained prior, which contains no field and no condition \\
Operating condition & Required: the velocity input carries the freestream direction and speed & Never given, in training or at test \\
Measurements & None & Any number of sensors, in one forward pass \\
Estimand & $\E[f\mid S,c]$ & $\E[f\mid S,y]$, with $c$ unreported \\
Pretraining data & ShapeNet cars, aircraft and watercraft: over 10,000 geometries, 1.35M samples & \nMeshes{} raw Objaverse-XL meshes \\
Backbone & Transolver, 3M to 15M parameters & Areal tokens, \nParams{} parameters \\
\bottomrule
\end{tabular}
\end{table}

\paragraph{No geometry model to benchmark.} GeoPT's network ``receives geometry and velocity as input during both
pre-training and fine-tuning'' \citep{geopt2026}. Its pretrained weights are a response of geometry to synthetic
dynamics, learned in one function, not a geometry model to which physics is added later. The question this paper asks,
whether a geometry model learned with no physics at all can serve as the prior for sensor-conditioned field models,
needs a geometry model that is separate from the physics. \areals{} has one; GeoPT has no geometry-only part that could
take its place under a sensor model.

\paragraph{A different estimand.} At fine-tuning, GeoPT replaces its random velocities with a task velocity that encodes
the simulation settings, for aerodynamics the freestream direction and speed, so it estimates $\E[f\mid S,c]$ and must
be told $c$. Every sensor benchmark here leaves $c$ unreported, so GeoPT has no input from which to read it, and it
takes no measurements. In our protocol its counterpart is the true-condition reference, which brackets the
sensor-conditioned models rather than competing with them. Making GeoPT read sensors would mean replacing its velocity
input with a sensor interface, which would produce a new model, not GeoPT.

\paragraph{Re-implementing GeoPT is not evaluating it.} GeoPT's trajectories need only geometry, so the \areals{} encoder
and decoder could be trained on them in place of surface completion. That run would test GeoPT's objective inside our
architecture, with our tokens, decoder and pretraining meshes, not the published model, its backbone or its corpus, so a
difference in either direction could not be credited to GeoPT or held against it. It would compare pretraining
objectives, a question this paper answers analytically for its own targets (\cref{sec:inversion}), rather than evaluate
GeoPT. The baselines for the sensor task are therefore those of \cref{sec:experiments}: the same transformer without
pretraining, GABI, explicit inference of the condition, and the shape-only and true-condition references.

\end{document}